\documentclass[10pt,twocolumn,letterpaper]{article}

\usepackage{iccv}              

\definecolor{iccvblue}{rgb}{0.21,0.49,0.74}
\usepackage[pagebackref,breaklinks,colorlinks,allcolors=iccvblue]{hyperref}

\def\paperID{*****} 
\def\confName{ICCV}
\def\confYear{2025}

\title{Motion Segmentation and Egomotion Estimation\\from Event‐Based Normal Flow}

\author{
  Zhiyuan Hua\thanks{* denotes equal contribution.},\quad
  Dehao Yuan\footnotemark[1],\quad
  Cornelia Fermüller \\[1ex]
  University of Maryland, College Park \\[0.5ex]
  \texttt{\{howardh,dhyuan,fermulcm\}@umd.edu}
}

\begin{document}
\maketitle
\begin{abstract}
This paper introduces a robust framework for motion segmentation and egomotion estimation using event-based normal flow, tailored specifically for neuromorphic vision sensors. In contrast to traditional methods that rely heavily on optical flow or explicit depth estimation, our approach exploits the sparse, high-temporal-resolution event data and incorporates geometric constraints between normal flow, scene structure, and inertial measurements. The proposed optimization-based pipeline iteratively performs event over-segmentation, isolates independently moving objects via residual analysis, and refines segmentations using hierarchical clustering informed by motion similarity and temporal consistency. Experimental results on the EVIMO2v2 dataset validate that our method achieves accurate segmentation and translational motion estimation without requiring full optical flow computation. This approach demonstrates significant advantages at object boundaries and offers considerable potential for scalable, real-time robotic and navigation applications.

\end{abstract}    
\section{Introduction}
\label{sec:intro}

Fundamental problems in visual motion understanding include the estimation of the sensor’s three-dimensional motion (egomotion) and the segmentation of independently moving objects. Solutions to these problems underpin higher-level navigation and manipulation tasks, such as localization, mapping, scene reconstruction, and object interaction.

Traditionally, both egomotion estimation and motion segmentation have relied on feature correspondences or optical flow. However, computing optical flow is computationally intensive and unreliable in certain image regions, particularly along object boundaries. Optical flow estimation requires at least two constraints. Local spatiotemporal information typically supports only the computation of a single component of motion, the so-called \textit{normal flow}, which lies along the image gradient direction. Recovering the second flow component generally requires additional assumptions about the smoothness of motion across the scene. Modern optical flow methods address this by incorporating multiple constraints such as temporal coherence, occlusion handling, and adaptive weighting.

This limitation has prompted researchers to question whether full optical flow is necessary even in the early days of motion analysis. Instead, could normal flow, derived entirely from local measurements, suffice for fundamental motion tasks? Various algorithms have since been developed to estimate 3D motion from normal flow \cite{BARRANCO2021,brodsky2000structure,fermuller1995passive,fermuller2013navigational,HUI2015422,negahdaripour1987direct,yuan2024learning}, and theoretical results have confirmed that egomotion estimation is indeed possible without full optical flow \cite{fermuller1998ambiguity,fermuller2000observability}. However, a complete solution for motion segmentation based solely on normal flow remains an open challenge.

This work is situated within the domain of neuromorphic vision. Neuromorphic engineering is a computing paradigm that draws inspiration from biological neural systems to design efficient hardware and algorithms. One of its most notable innovations is the event-based vision sensor \cite{lichtsteiner2008128}, which has garnered increasing attention in computer vision and robotics. Unlike conventional cameras that capture images at fixed frame rates, event-based sensors asynchronously record brightness changes at individual pixels, producing sparse, high-temporal-resolution data. These sensors offer significant advantages, including low power consumption, low latency, and high dynamic range, making them particularly well-suited for real-time, robust robotic perception \cite{gallego2020event}.

Given their sparse and asynchronous nature, event-based data are naturally aligned with normal flow estimation, which has become a compelling alternative to optical flow in neuromorphic perception. Benosman et al.~\cite{benosman2013event} first introduced a method for estimating normal flow from events by fitting planes to local spatiotemporal point clouds. Mueggler et al.~\cite{mueggler2015lifetime} later proposed a bio-inspired, causal version of this technique. However, the highly local nature of these methods results in limited accuracy, constraining their applicability in high-level vision tasks and preventing them from competing with modern optical flow techniques.

Recently, Yuan et al.~\cite{yuan2024learning} proposed a learning-based method for estimating normal flow from event data that achieves accuracy comparable to state-of-the-art optical flow algorithms. Their approach was later optimized for real-time execution \cite{yuan2025real}. Notably, this method performs especially well at object boundaries, where traditional optical flow methods often fail. The technique uses kernel-based methods to extract Random Fourier Features from local spatiotemporal neighborhoods. These features are then encoded into vectors that are input into a lightweight supervised neural network, which predicts the corresponding one-dimensional normal flow. This advancement establishes a strong foundation for developing bio-inspired, event-based solutions to visual motion interpretation.

A key challenge in motion analysis lies in the interdependence of 3D motion estimation and scene segmentation—often described as a chicken-and-egg problem. Accurate estimation of the sensor's 3D motion requires knowledge of the static background, free from the influence of independently moving objects. Conversely, reliable segmentation of independently moving objects often depends on knowledge of the underlying 3D motion.

In this paper, we propose a classical optimization-based framework for the joint estimation of 3D sensor motion, segmentation of independently moving objects, and estimation of their respective motions. Our approach uses as input the estimated normal flow and rotational measurements from an inertial measurement unit (IMU). The method processes data in discrete event slices and proceeds iteratively. For each slice, it begins by fitting a simple planar rigid motion model to the background, yielding an initial segmentation. It then refines this segmentation using 3D motion estimates from the previous slice by jointly tracking both the background and the moving objects. This refinement step combines clustering based on normal flow with motion-based background tracking. Finally, the 3D motions of both the sensor and the objects are re-estimated, as detailed in Section~\ref{sec:methodology}.

\section{Related Work}
\subsection{Feature Tracking and SLAM.} 
In the early stages of event-based egomotion estimation, feature tracking and SLAM were the dominant approaches in the field. The common methodology of feature tracking and SLAM with event-based vision systems is to exploit the asynchronous, high-temporal-resolution nature of event streams for continuous pose estimation and map construction. As two representative examples, \cite{kim2016real} uses a probabilistic filtering framework that decouples the estimation of camera pose, scene gradients, and depth, enabling real-time 3D reconstruction and 6-DoF tracking. \cite{rebecq2016evo} adopts a geometric approach that aligns events with a semi-dense 3D model using image-to-model tracking, achieving high-frequency pose estimation even under challenging conditions. There are many derivatives from this mainstreaming methodology. For example, probabilistic and filtering-based tracking \cite{weikersdorfer2012event,delbruck2007fast,mueggler2015towards}, geometric and direct tracking, \cite{rebecq2016evo,bryner2019event,kim2008simultaneous}, inertial sensing integration for improved robustness \cite{rebecq2017real,kueng2016low}, full SLAM systems for navigation and exploration \cite{weikersdorfer2013simultaneous,milford2015towards,liu2020ground}, and low-latency reactive control in robotics \cite{delbruck2013robotic,censi2013low}.


\subsection{Learning-Based Approaches}
More recently, with the emergence of event camera datasets \cite{mitrokhin2019ev,burner2022evimo2} and advances in deep learning, many approaches have been proposed for learning-based motion segmentation and egomotion estimation. These approaches utilize various neural network architectures, including convolutional networks \cite{sanket2020evdodgenet,zhu2019unsupervised,jiang2024event,chen2025supereio,klenk2024deep,tabia2022deep}, recurrent networks \cite{zhang2023multi,zhu2024continuous,guan2024deio,nguyen2019real}, attention-based networks \cite{lin20226,arja2024motion,alkendi2024neuromorphic,georgoulis2024out,zhou2024event,messikommer2023data}, spiking neural network \cite{zheng2022spike,kirkland2020spikeseg,nagaraj2023dotie}, graph neural network \cite{mitrokhin2020learning}, and implicit neural representation \cite{ma2024continuous}. The tasks are typically categorized as motion segmentation only \cite{zhang2023multi,jiang2024event,zhu2024continuous,arja2024motion}, egomotion estimation only \cite{zhu2019unsupervised,zheng2022spike,lin20226}, and both \cite{mitrokhin2019ev}. Training strategies are generally divided into supervised learning and unsupervised learning \cite{zhu2019unsupervised,zheng2022spike,arja2024motion,chen2025supereio}. Although these methods perform well on benchmark datasets, they often suffer from significant performance degradation when applied to data from different domains. This is primarily because the networks tend to overfit to the specific characteristics of the training scenes.


\subsection{Contrast Maximization and Optical Flow}
To improve the robustness of egomotion estimation and motion segmentation, recent works explore using the intermediate computation of optical flow or contrast maximization to enhance the computation. Building on the optical flow \cite{conradt2015board,stoffregen2018simultaneous,serres2016event,pijnacker2018vertical,raharijaona2017toward,abdul2016estimating} or contrast maximization \cite{zhou2021event,gallego2017accurate,parameshwara20210,huang2023progressive,huang2023mc,gallego2019focus,stoffregen2019event,gallego2018unifying}, these methods define geometric constraints and solve the egomotion and motion segmentation by optimization. Contrast maximization (CM) approaches are usually more accurate than optical flow approaches, but solving contrast maximization is typically expensive. Optical flow enables faster solving of egomotion and motion segmentation, but estimating optical flow robustly across different domains is still a challenging problem.
\subsection{Event-based Normal Flow}
A few works have also used normal flow for ego-motion estimation. For example, Lu et al.\cite{lu2023event} address the linear velocity estimation task for drones under aggressive maneuvers in a stereo setting, utilizing stereo and IMU data. Ren
et al. \cite{ren2024motion} established specific constraints between instantaneous motion-and-structure
parameters and event-based normal flow for ego-motion and depth estimation. Yuan et
al. \cite{yuan2024learning}] developed a robust estimator for the direction of translation from events and IMU, and in \cite{yuan2025real} presented a real-time implementation. However, event-based normal flow has not yet been used for segmentation.
\section{Methodology}
\label{sec:methodology}

\begin{figure*}[ht]
  \centering
  \includegraphics[width=0.95\linewidth]{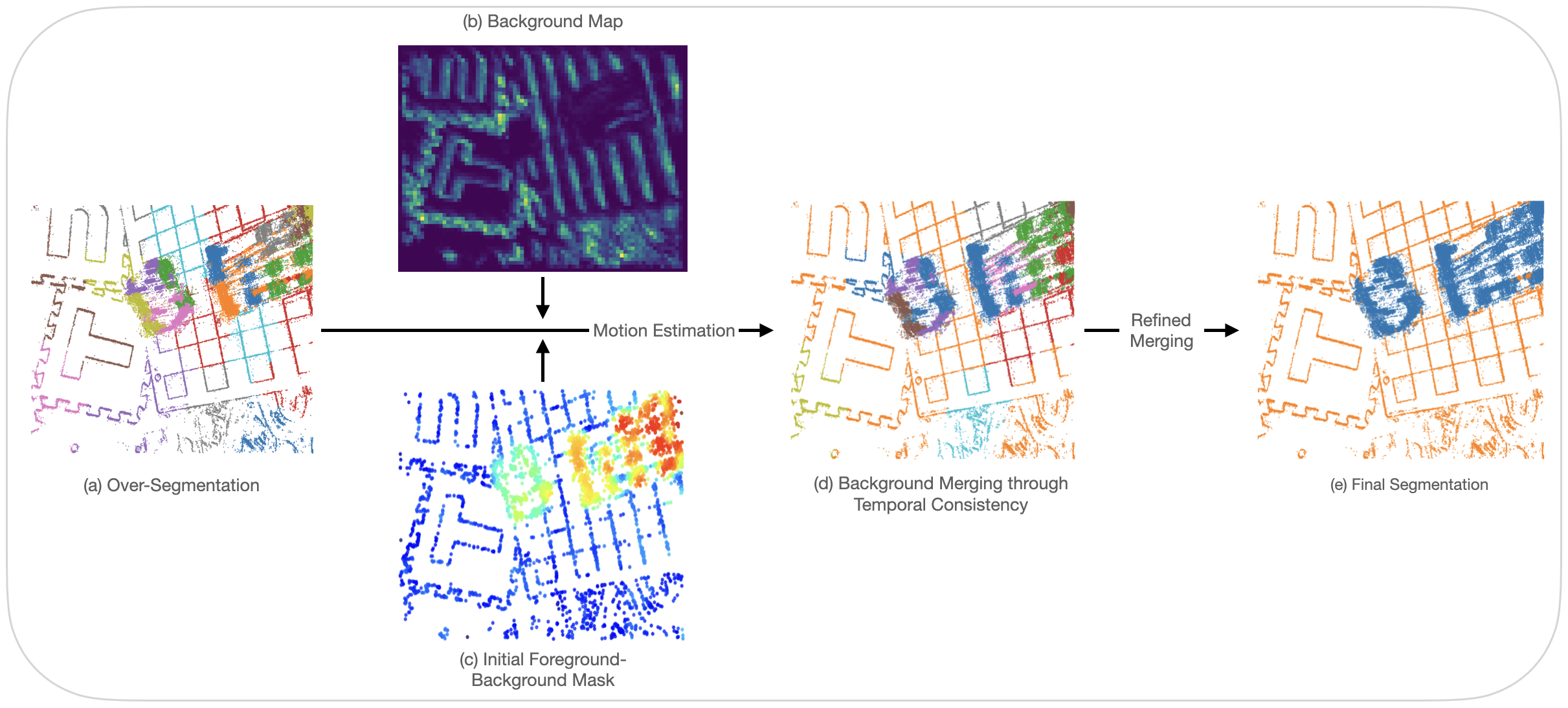}
  \caption{Overview of our motion segmentation pipeline. 
    \textbf{(a)} The pipeline begins with an over-segmentation of event data based on spatial proximity and normal flow orientation. 
    \textbf{(b)} A map of background events is maintained using an exponential moving average across frames. 
    \textbf{(c)} Initial foreground-background separation is performed by clustering normal flow residuals. 
    Combining the priors from \textbf{(a)}\textbf{(b)}\textbf{(c)}, the motions of the background are estimated for the current frame.
    \textbf{(d)} Using estimated motion model to warp prior background, new background clusters are merged based on temporal consistency.
    \textbf{(e)} Final motion-based segmentation is produced via hierarchical refined merging using motion similarity and residual coherence.}
  \label{fig:pipeline_only}
\end{figure*}

Our approach aims to segment independently moving objects and estimate camera and object motion from event-based normal flow. The core idea is to leverage the geometric constraints between normal flow, 3D motion, and scene structure. To separate the background from the foreground, the method iteratively solves and refines the solution, thereby increasing the robustness. By initially modeling the background with a planar assumption, we identify deviations caused by moving objects as residuals. These residuals serve as a cue for segmentation, which is then refined through temporal consistency and 3D motion similarity. This unified formulation allows robust motion segmentation without explicit depth or optical flow estimation.

\subsection{Problem Setting and Algorithm Overview}
Our motion segmentation and ego-motion estimation algorithm is recursive. We assume at a specific step, we have the following inputs:
\begin{enumerate}
    \item slice of events at the \textcolor{blue}{current} step 
    \(\mathcal{E}' = \{(t_j, x_j, y_j)\}_{j=1}^{n'}\),
    \item normal flow vectors estimated for each event at the \textcolor{blue}{current} step,
    \item background mask \(\mathcal{M}_0\) and background motion parameters \(\mathcal{P}_0\) at the \textcolor{green}{previous} step,
    \item motion segmentation masks and motion parameters of each segment at the \textcolor{green}{previous} step \((\mathcal{P}_1, \mathcal{P}_2, \ldots, \mathcal{P}_N)\).
\end{enumerate}
We compute the following outputs at each recursive step:
\begin{enumerate}
    \item number of motion segments at the \textcolor{blue}{current} step \(N'\),
    \item motion segmentation masks at the \textcolor{blue}{current} step \((\mathcal{M}_1, \mathcal{M}_2, \ldots, \mathcal{M}_{N'})\) that are consistent with the previous step\footnote{Since moving objects may emerge, persist or vanish, we create new masks for emerging objects, match masks for persisting objects, and delete masks for vanishing objects.}, and motion parameters of each segment at the \textcolor{blue}{current} step \((\mathcal{P}_1, \mathcal{P}_2, \ldots, \mathcal{P}_{N'})\),
    \item background mask \(\mathcal{M}_0'\) and background motion parameters \(\mathcal{P}_0'\) at the \textcolor{blue}{current} step.
\end{enumerate}
As a result, our pipeline outputs three time series: motion segmentation masks, segment-wise motion parameters, and egomotion estimation.

\textbf{Algorithm Overview}
The recursive algorithm begins by computing per-event normal flow using a pre-trained estimator \cite{yuan2025real, yuan2024learning}. It then proceeds through four stages (see Fig.~\ref{fig:pipeline_only}:
\begin{enumerate}
    \item \textbf{Initial Normal Flow Clustering (Scene Over-Segmentation)} (Sec.~\ref{sec:over-segmentation}): perform k-means clustering on the normal flow and image coordinates to generate a coarse segmentation of the current event slice. The number of clusters is set higher than the actual number of objects to ensure over-segmentation. This clustering feeds into stages 3 and 4.
    \item \textbf{Preliminary Foreground–Background Segregation} (Sec.~\ref{sec:preliminary-segmentation}): using IMU rotation data and normal flow, fit a simple 3D motion model (planar scene) to coarsely segregate the scene, identifying potential independently moving objects (IMOs) through residual analysis. This provides an initial foreground-background mask.
    \item \textbf{Coarse Segment Merging through Temporal Consistency} (Sec.~\ref{sec:pre-merge}): estimate the 3D background motion from normal flow, warp the previous step’s background mask to the current step, and merge coarse segments from stage 1 based on temporal consistency.
    \item \textbf{Refined Segment Merging through Motion Similarity} (Sec.~\ref{sec:refined-merge}): iteratively refine the segmentation by merging segments whose fitted 3D object motions are within a predefined similarity threshold, stopping when convergence is reached.
\end{enumerate}

\subsection{Initial Normal Flow Clustering}
\label{sec:over-segmentation}

Due to the large number of events within each slice, clustering them directly based on motion is computationally prohibitive. To address this, we first apply k-means clustering to reduce the problem to a manageable size. Each event is represented by a feature vector consisting of its pixel coordinates $x_j, y_j$ and normal flow $n_j$:
\begin{equation}
    \left[x_j\:y_j\:\lambda n_j \right]
\end{equation}
This over-segmentation strategy is inspired by \cite{yuan2024learning}, which demonstrates that normal flow predictions preserve the boundaries of moving objects effectively. We set the number of clusters to 30 and use a weighting factor $\lambda = 0.5$ to balance spatial and motion information. By intentionally over-segmenting the events, we ensure that those with clearly similar motion patterns are grouped together.

\begin{figure}[h]
    \centering
    \includegraphics[width=\linewidth]{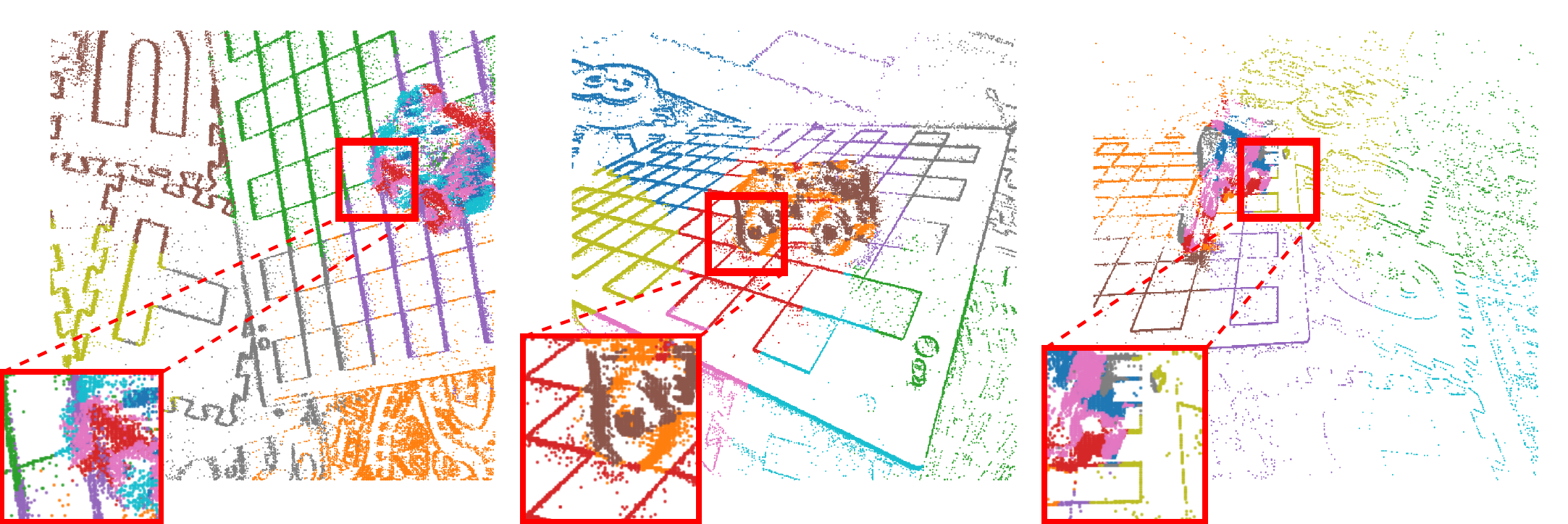}
    \caption{Visualization of the over-segmentation results. Although a moving object may be divided into multiple segments, its boundaries are well preserved, providing a strong initialization for the subsequent refinement process. The event colors indicate the cluster assignments.}
    \label{fig:over-seg}
\end{figure}

The underlying intuition is that if two events are close in both pixel coordinates and normal flow, they are highly likely to belong to the same motion segment. However, events with dissimilar features may still share the same motion, and such cases will be handled in later refinement stages. Figure \ref{fig:over-seg}(a) illustrates examples of the initial over-segmentation: while object boundaries are sharply preserved, individual moving objects may be divided into multiple clusters, which will subsequently be merged. These clusters serve as input to the later refinement segmentation stages (Sections \ref{sec:pre-merge}, \ref{sec:refined-merge}).

\subsection{Preliminary Foreground--Background Segregation}
\label{sec:preliminary-segmentation}

Prior to the main segmentation stages, we perform a preliminary coarse segmentation to distinguish between the background and potential independently moving objects (IMOs). This step utilizes the event-based normal flow $\mathbf{n_j}$ and IMU-provided rotation velocity $\mathbf{w}$ to estimate the camera's 3D motion and the scene's depth structure. We model the scene as a general plane, fitting it to the normal flow by solving for the eight parameters $\mathbf{a} = (a_1 \ldots a_8)^{T}$ that characterize the 3D motion and plane parameters (a homography).

In some more detail, let us denote the optical flow as $\mathbf{u}$, the normal flow as $\mathbf{n}$, with $\mathbf{n_0}$ a unit vector in the direction of the normal flow and $n$ the length of the normal flow, and $n = \mathbf{u}^T \mathbf{n_0}$. The equations relating flow to 3D motion (with rotation $\mathbf{w} = (\omega_x, \omega_y, \omega_z)^T$ and translation $\mathbf{t}= (t_x, t_y, t_z)^T$) and the depth $Z(\mathbf{x})$ at a point $\mathbf{x} = (x,y)$ are written as:
\begin{equation}
 \mathbf{u} = \Bigl(\frac{1}{Z} A(\mathbf{x}) \mathbf{t} + B(\mathbf{x}) \mathbf{w}\Bigr)
 \label{eq:flow_ego}
\end{equation}
Thus, the normal flow amounts to:
\begin{equation}
u_n(\mathbf{x}) = \mathbf{u}(\mathbf{x})^T \mathbf{n_0}(\mathbf{x}) = \Bigl(\frac{1}{Z} A(\mathbf{x}) \mathbf{t} + B(\mathbf{x}) \mathbf{w}\Bigr)^T \mathbf{n_0}
\label{eq:normal_ego}
\end{equation}
with 
\begin{equation}
 A(\mathbf{x}) = \begin{bmatrix}
-1 & 0 & x\\
0 & -1 & y
\end{bmatrix}
\end{equation}
and 
\begin{equation}
 B(\mathbf{x}) = \begin{bmatrix}
x y & -(1 + x^2) & y\\
1 + y^2 & -x y & -x
\end{bmatrix}
\end{equation}

If we assume the scene in view to be a plane, the depth $Z(\mathbf{x})$ at a point can be expressed as $\tfrac{d}{Z(\mathbf{x})} = \alpha x + \beta y + \gamma$, with $(\alpha, \beta, \gamma)^T$ the surface normal vector to the plane, and $d$ the distance of the plane to the origin. Then Equation~(\ref{eq:normal_ego}) becomes:
\begin{equation}
u_n(\mathbf{x}) = \bigl(C(\mathbf{x}) \mathbf{a}\bigr)^T \mathbf{n_0} = \bigl(C(\mathbf{x})\mathbf{n_0}\bigr)^T \mathbf{a}
\label{eq:normal_plane}
\end{equation}
with 
\begin{equation}
 C(\mathbf{x}) = \begin{bmatrix}
x^2 & x y & x& y & 1& 0 & 0& 0\\
x y & y^2 &0 &0&0& y& x& 1
\end{bmatrix}
\end{equation}
and 
\begin{equation}
 \mathbf{a} = \begin{bmatrix}
- d \omega_y + t_z \alpha \\
d \omega_x + t_z \beta\\
t_z \gamma - t_x \alpha\\
d \omega_z + t_x \beta\\
-d \omega_y - t_x \gamma \\
t_z \gamma - t_y \beta\\
-d \omega_z - t_y \alpha\\
d \omega_x - t_y \gamma
\end{bmatrix}
\end{equation}

Equation~(\ref{eq:normal_plane}) imposes a constraint for each normal flow measurement. By aggregating these constraints across all events, we formulate a linear system and solve for $\mathbf{a}$ using least squares. Residuals are computed as the differences between the observed and predicted normal flow values. Events with large residuals are identified as potential independently moving objects (IMOs) through clustering. The resulting mask provides an initial, coarse separation of foreground and background, which serves as a crucial input to the subsequent over-segmentation and region merging stages of our pipeline.

\paragraph{Initial Background Cluster Assignment via Residuals}

To robustly segment independently moving objects (IMOs), we apply K-Means clustering on the residual magnitudes obtained from the least squares fitting. The key idea is that pixels associated with IMOs typically exhibit significantly higher residuals compared to the background, making clustering a viable strategy for coarse separation. After computing residual magnitudes for all events, we first smooth the residuals using a Gaussian filter in the image grid space. This denoising step, implemented via a grid-based, efficient convolution using a Gaussian kernel, helps reduce noise and improves the robustness of subsequent clustering. To assign semantic meaning to the clusters, we analyze the cluster centers: the one with the lower average residual is heuristically assumed to represent the background, while the other represents potential IMOs. This initial background assignment helps us identify background where we don't have prior knowledge from a previous frame, or when we need to re-initialize the background assignment.

\subsection{Coarse Segment Merging through Temporal Consistency}
\label{sec:pre-merge}

\paragraph{Initialization}

During the first few frames, where no historical background data is available, we rely on the residual-based segmentation to initialize a coarse background mask. This serves as a proxy until temporal cues become available.

\paragraph{Background Warping and Matching}

At each new frame, the coarse segmentation from Sec.~\ref{sec:preliminary-segmentation} is refined in the background region by leveraging temporal consistency across frames, utilizing previous motion parameters and background information. We assume access to the previous background mask $\mathcal{M}_0$, represented by event coordinates $\mathbf{x}_{\text{prev}}$, and 3D motion parameters $(\mathbf{t}_{\text{prev}}, \mathbf{w}_{\text{prev}})$.

We warp $\mathbf{x}_{\text{prev}}$ to the current frame using the optic flow displacement (described in Eq.~\ref{eq:flow_ego}) derived from previous motion. We simplify by using a planar motion model with constant depth.  

Warped points $\mathbf{x}_{\text{warped}}$ are matched to current event coordinates $\mathbf{x}$ using appearance-based matching for each cluster, identifying this way the background pixels that belong to the matching background clusters, and we index them as $\mathcal{I}_{\text{bg}}$. 

\paragraph{Using Refined Background Motion to Improve Coarse Merging}

To refine background motion, using the coarsely identified background pixels, $\mathbf{x}[\mathcal{I}_{\text{bg}}]$, we estimate the current translation velocity $\mathbf{t}_{\text{est}}$ from the corresponding normal flow vectors $\mathbf{n}[\mathcal{I}_{\text{bg}}]$. IMU provides the rotation.

This involves solving a linear Support Vector Machine (SVM) \cite{yuan2024learning} classification problem using as input the derotated normal flow (i.e.\ $\mathbf{n}_{\text{derot}} = \mathbf{n} - (B(\mathbf{x}) \mathbf{w})^T \mathbf{n_0}$). The estimated $\mathbf{t}_{\text{new}}$ refines the background hypothesis, and the corresponding normal flow residuals $\mathbf{r}_{\text{new}} = \mathbf{n} - \mathbf{n}(\mathbf{x}, \mathbf{t}_{\text{new}}, \mathbf{w})$ are computed for use in merging. 

This refined background motion helps improve the warping, and further merge clusters that belong to the background at this stage. It is crucial to pre-emptively merge as many background clusters in parallel using temporal knowledge to greatly improve efficiency and reduce the more costly sequential fine merging operations needed in the next step.

\paragraph{Background Identification}

To ensure that the semantic background assignment is consistent and robust across frames, the background identification prioritizes shape similarity between the coarsely merged background and a persistent background map $\mathcal{M}_0'$, updated via an exponential moving average (EMA) with adaptive $\alpha$ based on similarity, ensuring temporal consistency. The persistent background map was accumulated from all previous frames' events that were classified as background; the exponential moving average ensures that we retain a background map even if we don't have successful background separation in some frames.

Next, we describe the merging of background regions obtained in the over-segmentation (in Section~\ref{sec:preliminary-segmentation}).

\subsection{Hierarchical Segment Merging through Refinement}
\label{sec:refined-merge}

After the previous two stages, we have a preliminary segmentation that is based on normal flow and temporal consistency. In this stage, we merge the segments hierarchically to obtain the final refined segmentation. The refinement is based on the classical hierarchical clustering algorithm \cite{bridges1966hierarchical}. 

For each cluster in the image (obtained in Sec.~\ref{sec:preliminary-segmentation}), we estimate a 3D translation assuming a planar shape model. Specifically, we solve for $\mathbf{t}$ within each patch, setting $Z = 1$ using a Tikhonov-regularized least squares formulation, i.e.,
\[
\sum_{i=1}^N \bigl(A(\mathbf{x}_i)^\top A(\mathbf{x}_i) + \lambda I\bigr) \,\mathbf{t}
\;=\;
\sum_{i=1}^N A(\mathbf{x}_i)^\top \mathbf{n}_{\text{derot}}(\mathbf{x}_i),
\]
where $\lambda = 10^{-6}$ is the regularization parameter and $I$ is the $3\times3$ identity matrix.

The final stage consolidates clusters into coherent motion segments by iteratively merging those with similar motion patterns. For each pair of active clusters $C_i$ and $C_j$, we compute a similarity score:
\begin{equation}
    \text{Similarity}(C_i, C_j) = -\frac{\|\mathbf{t}_i - \mathbf{t}_j\|^2}{\|\mathbf{t}_i\|^2 + \|\mathbf{t}_j\|^2}
    \;-\;\lambda_r \|\bar{r}_i - \bar{r}_j\|^2
\end{equation}
where:
\begin{itemize}
    \item $\mathbf{t}_i$ and $\mathbf{t}_j$ are estimated translations for clusters $C_i$ and $C_j$.
    \item $\bar{r}_i$ and $\bar{r}_j$ are the \textit{mean per-event residuals} within each cluster.
    \item $\lambda_r = 0.5$ is a weighting factor that balances motion similarity and residual consistency.
\end{itemize}

If available, the residuals $\bar{r}_i$ and $\bar{r}_j$ are computed using the updated residuals $r_{\text{new}}$ from Sec.~\ref{sec:pre-merge}, which incorporate refined background motion. Otherwise, the original residuals are used. A penalty is added if one cluster matches the persistent background map while the other does not, preventing background–foreground merges.

Importantly, merging is only allowed between spatially connected clusters, i.e., clusters whose bounding boxes overlap in image space. This ensures that disjoint regions with similar motion (e.g., separate hands or objects) are not mistakenly fused.

To avoid erroneous merges between background and foreground segments, we introduce a penalty when only one cluster matches the persistent background map. This additional term discourages merging across background–foreground boundaries based on histogram similarity to the persistent background descriptor $\mathcal{M}_0'$.

Merging proceeds by selecting the pair with the highest similarity above a threshold, combining their event indices, and recomputing $\mathbf{t}$ and centroids. The process iterates until no pair exceeds the threshold or fewer than two clusters remain.

The resulting merged clusters are tracked using a Kalman Filter–based tracker, which assigns each segment a consistent track ID (ID 0 for background, positive integers for foreground). Final segment labels are derived from these assignments, ensuring motion-based temporal coherence and completing the hierarchical refinement process.

A new, refined estimation of the background motion model on the now-refined background is then solved using the linear SVM discussed in Sec.~\ref{sec:pre-merge}, to be used in the next frame.

\section{Experiments}
\label{sec:experiment}

This section quantitatively and qualitatively evaluates our method on the EVIMO2v2 dataset \cite{burner2022evimo2}, focusing on Intersection over Union (IoU) for segmentation accuracy and Root Mean Square Error (RMSE) for translational motion estimation. We present quantitative results and visual comparisons, highlighting the method’s performance in segmenting moving objects and estimating camera and object motion across diverse scenes.

\subsection{Intersection over Union (IoU)}
\begin{figure}[ht]
\centering
\setlength{\tabcolsep}{0pt}

\begin{tabular}{ccccc}
\includegraphics[width=0.2\linewidth]{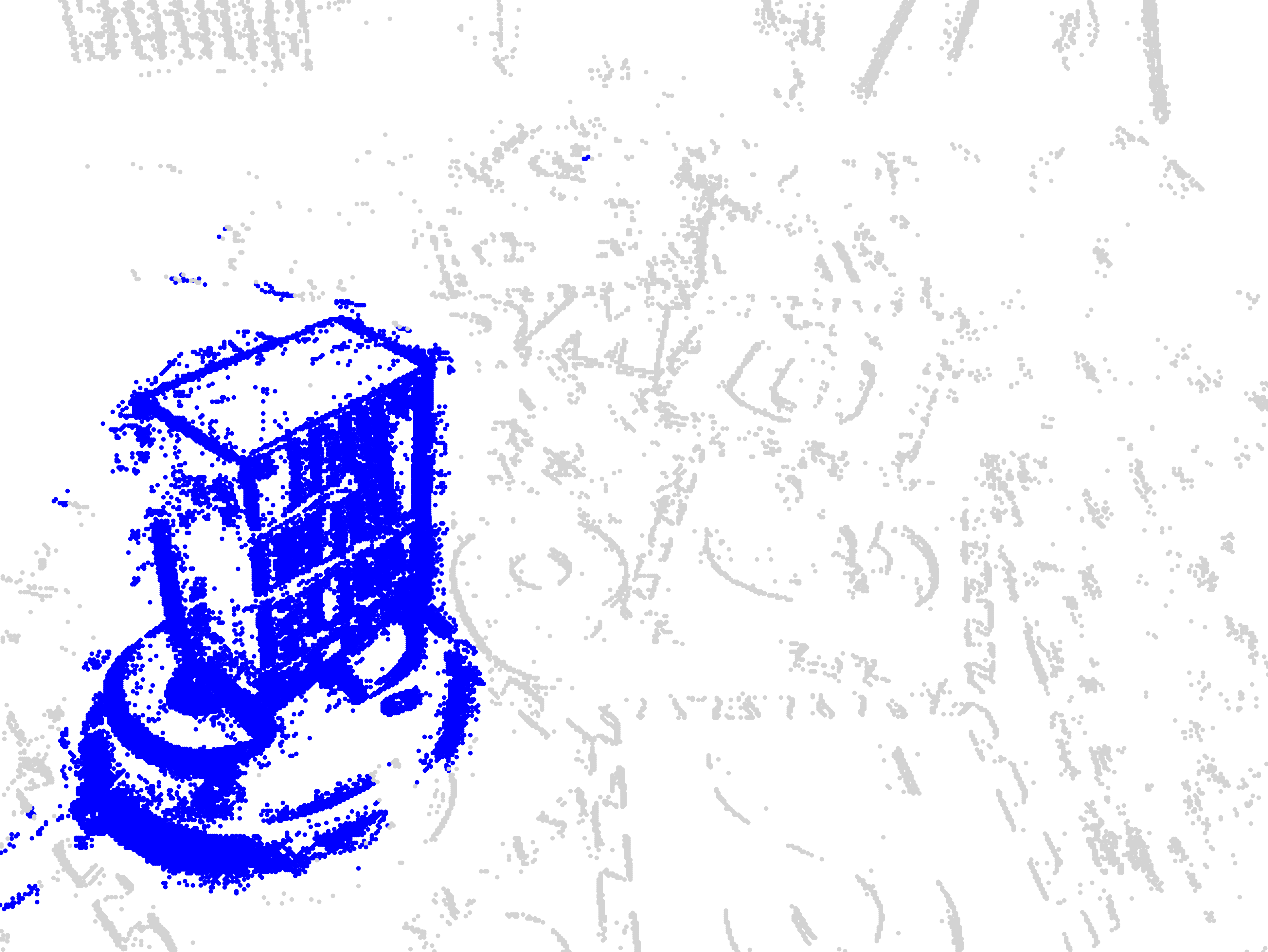} &
\includegraphics[width=0.2\linewidth]{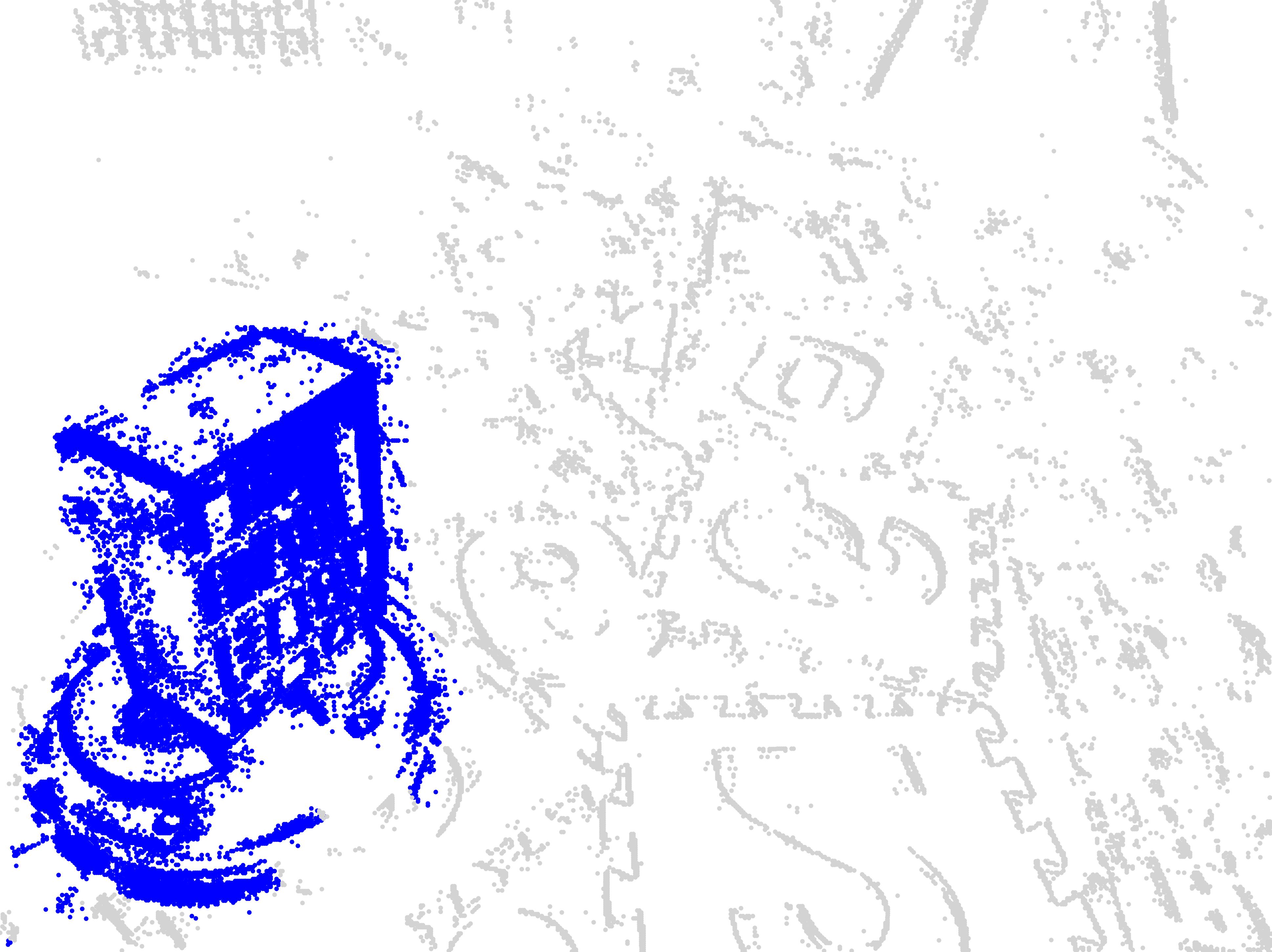} &
\includegraphics[width=0.2\linewidth]{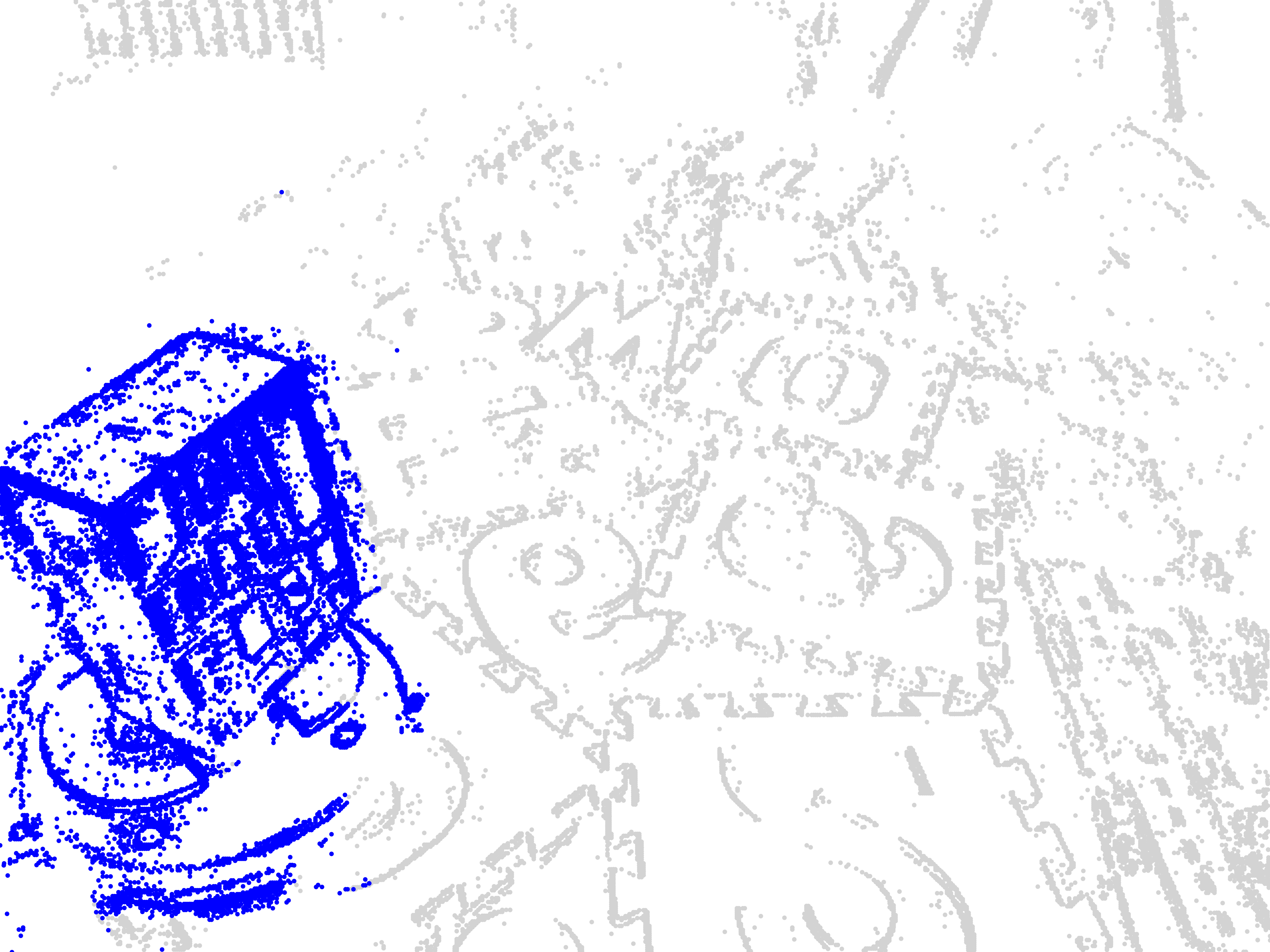} &
\includegraphics[width=0.2\linewidth]{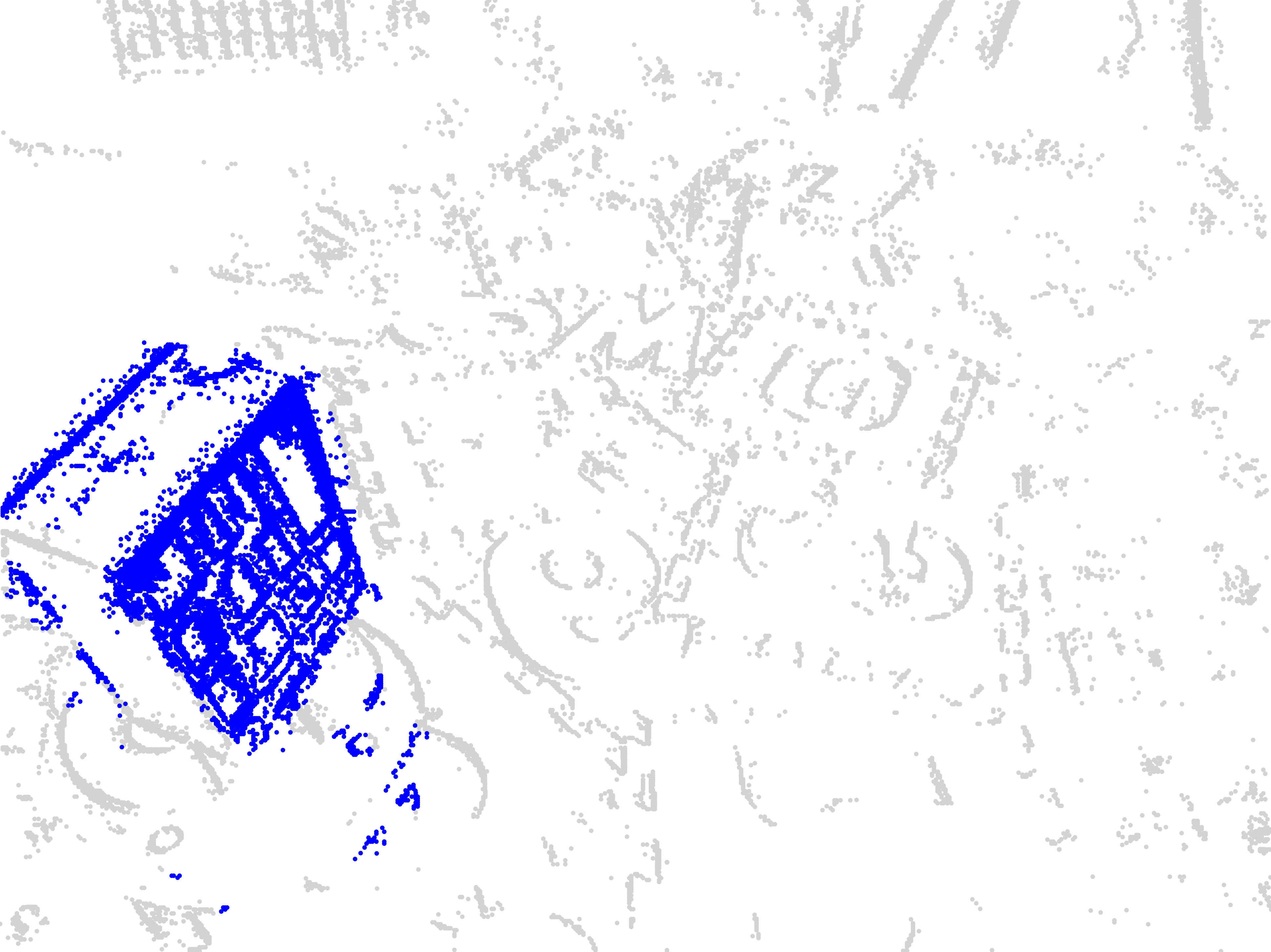} &
\includegraphics[width=0.2\linewidth]{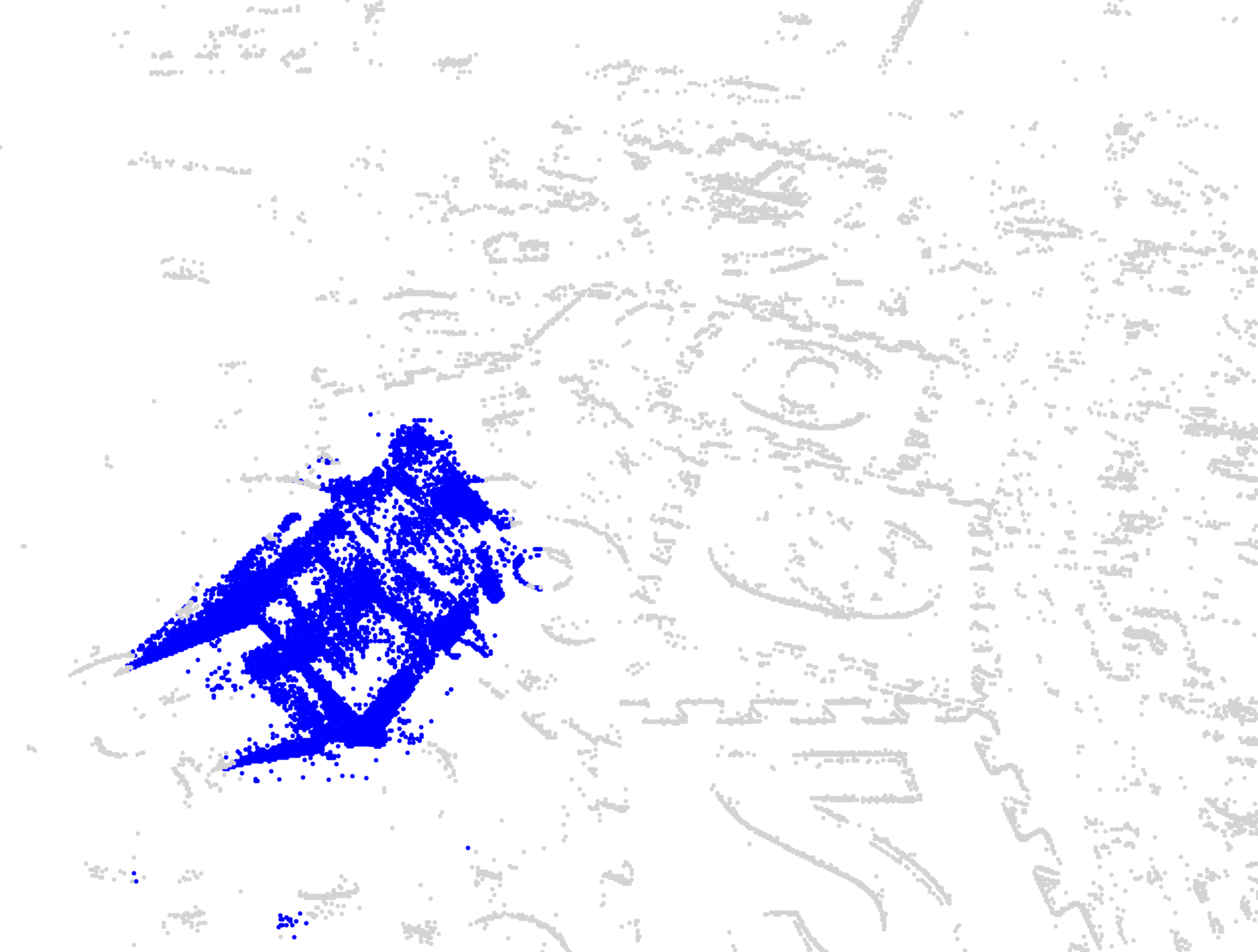} \\
 0.809 &  0.870 &  0.884 & 0.783 & 0.903 \\
\end{tabular}
\caption*{Frames from sequence 13-00}
\vspace{2mm}
\begin{tabular}{ccccc}
\includegraphics[width=0.2\linewidth]{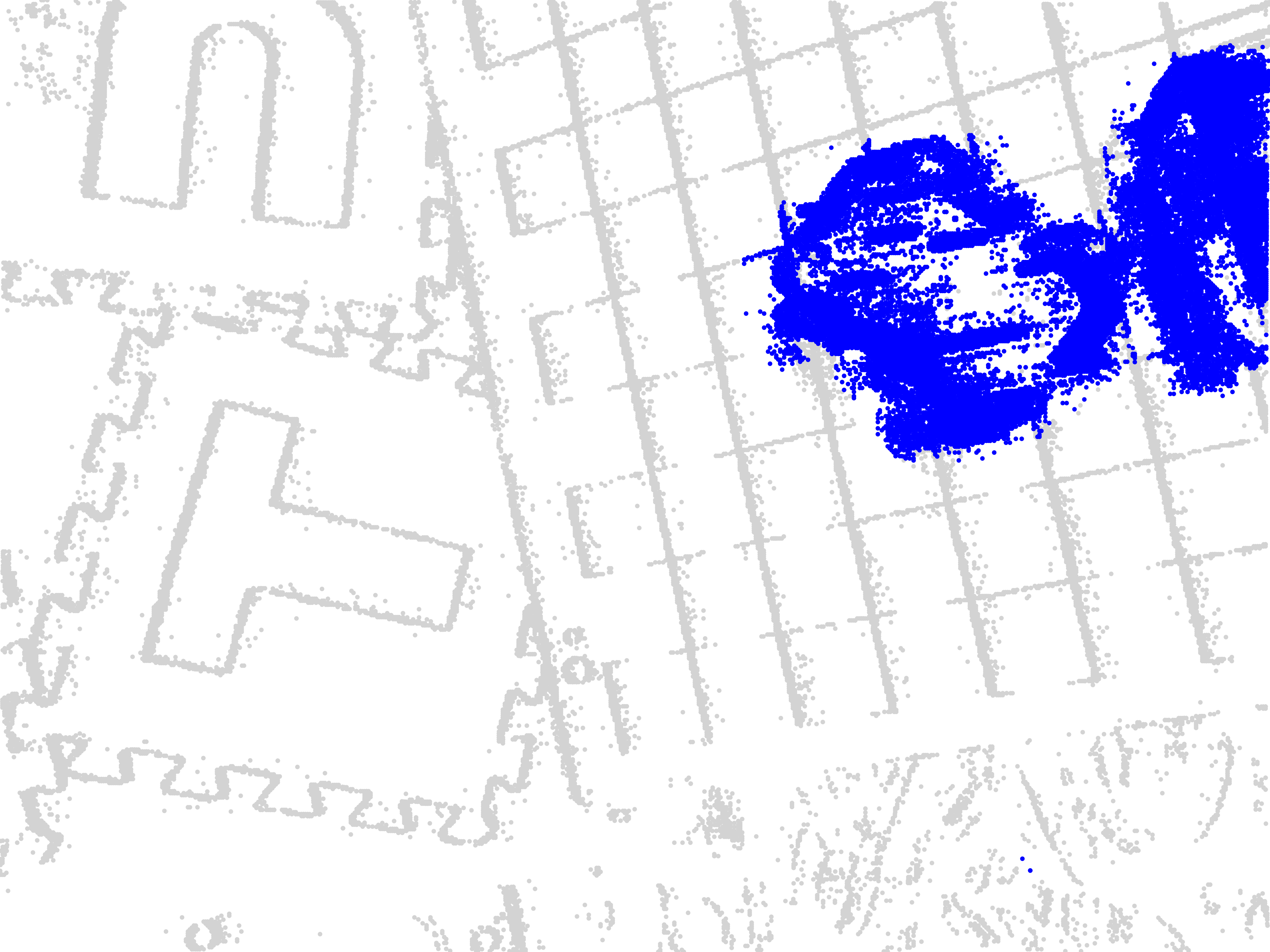} &
\includegraphics[width=0.2\linewidth]{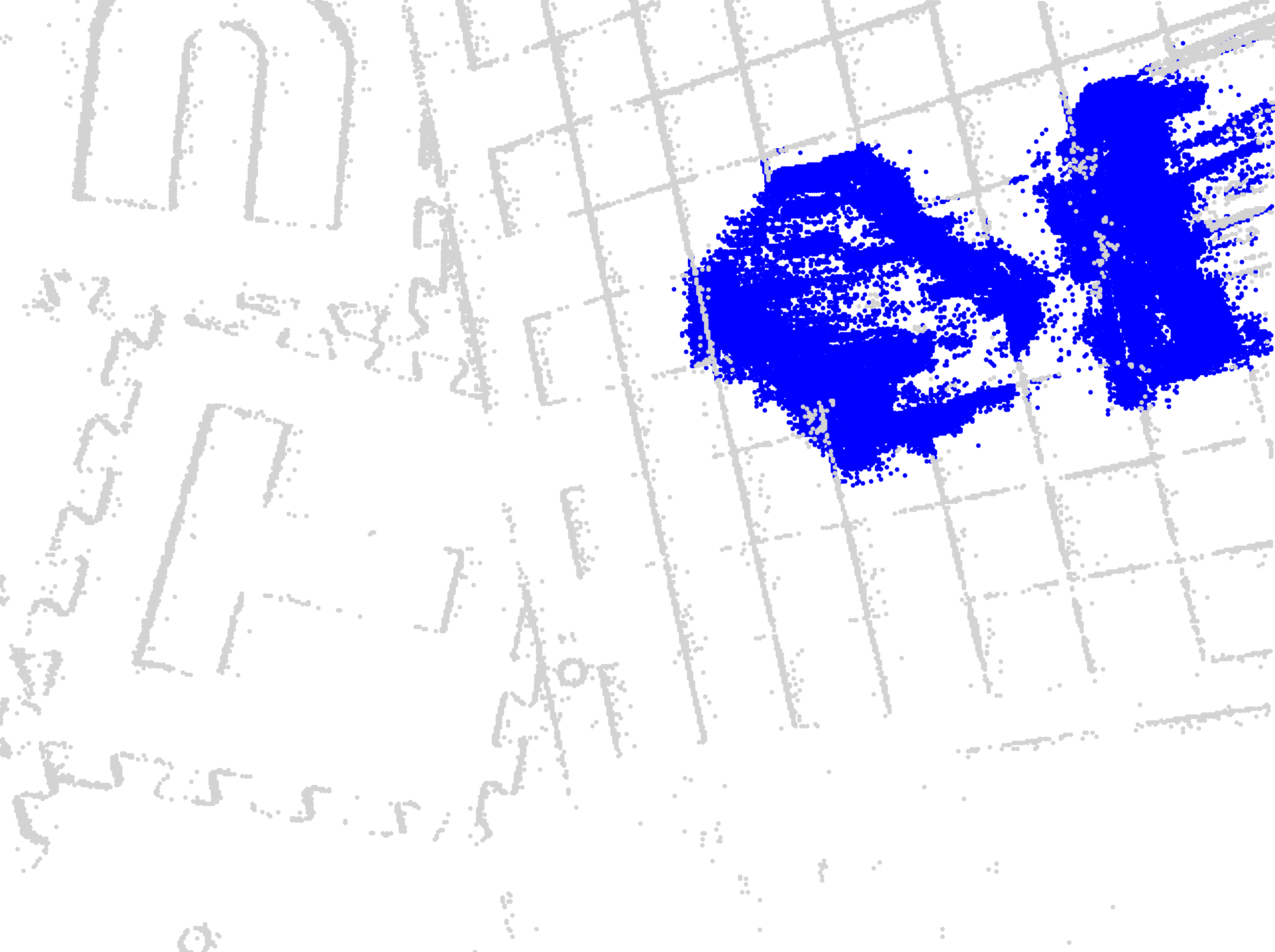} &
\includegraphics[width=0.2\linewidth]{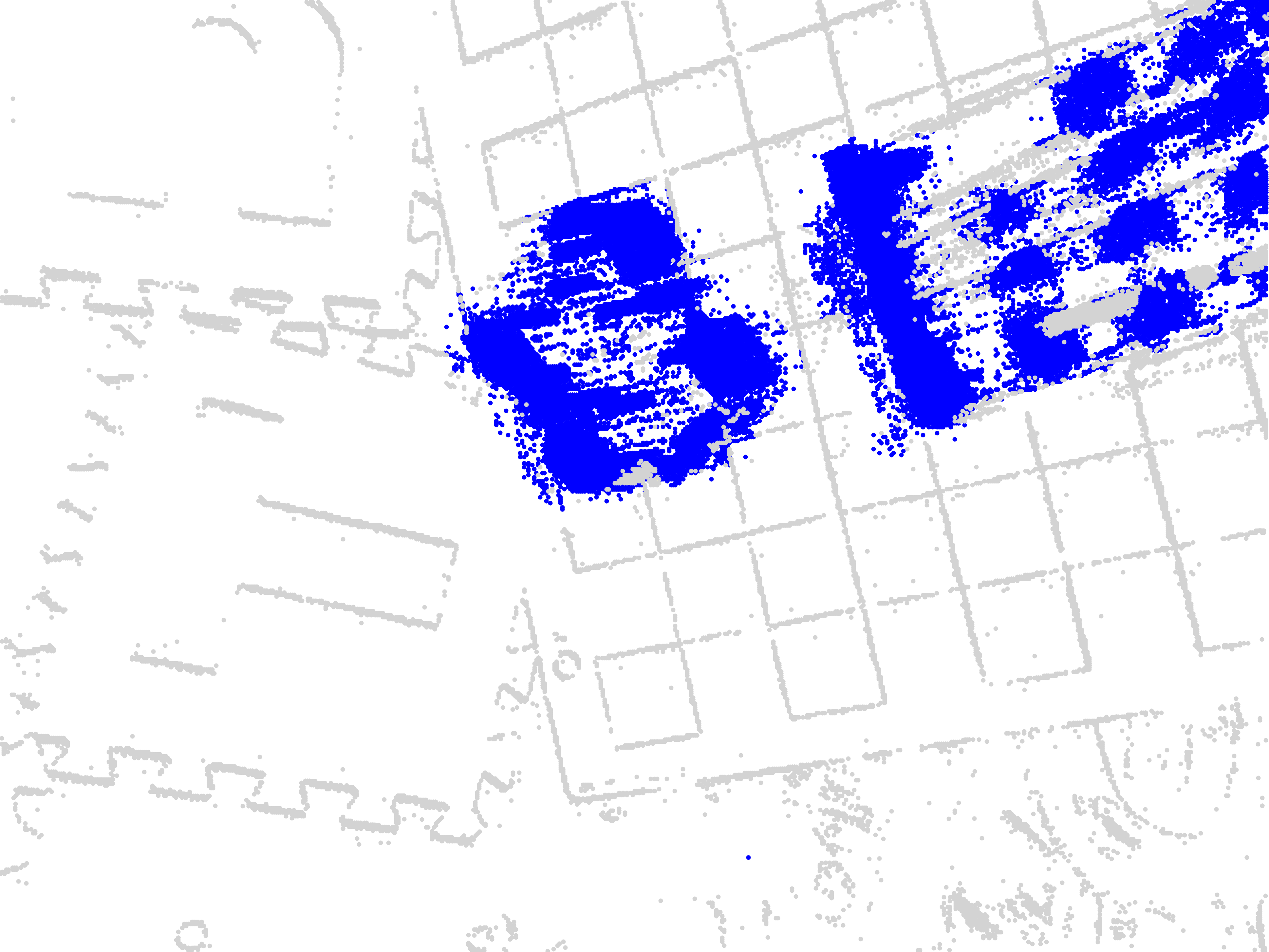} &
\includegraphics[width=0.2\linewidth]{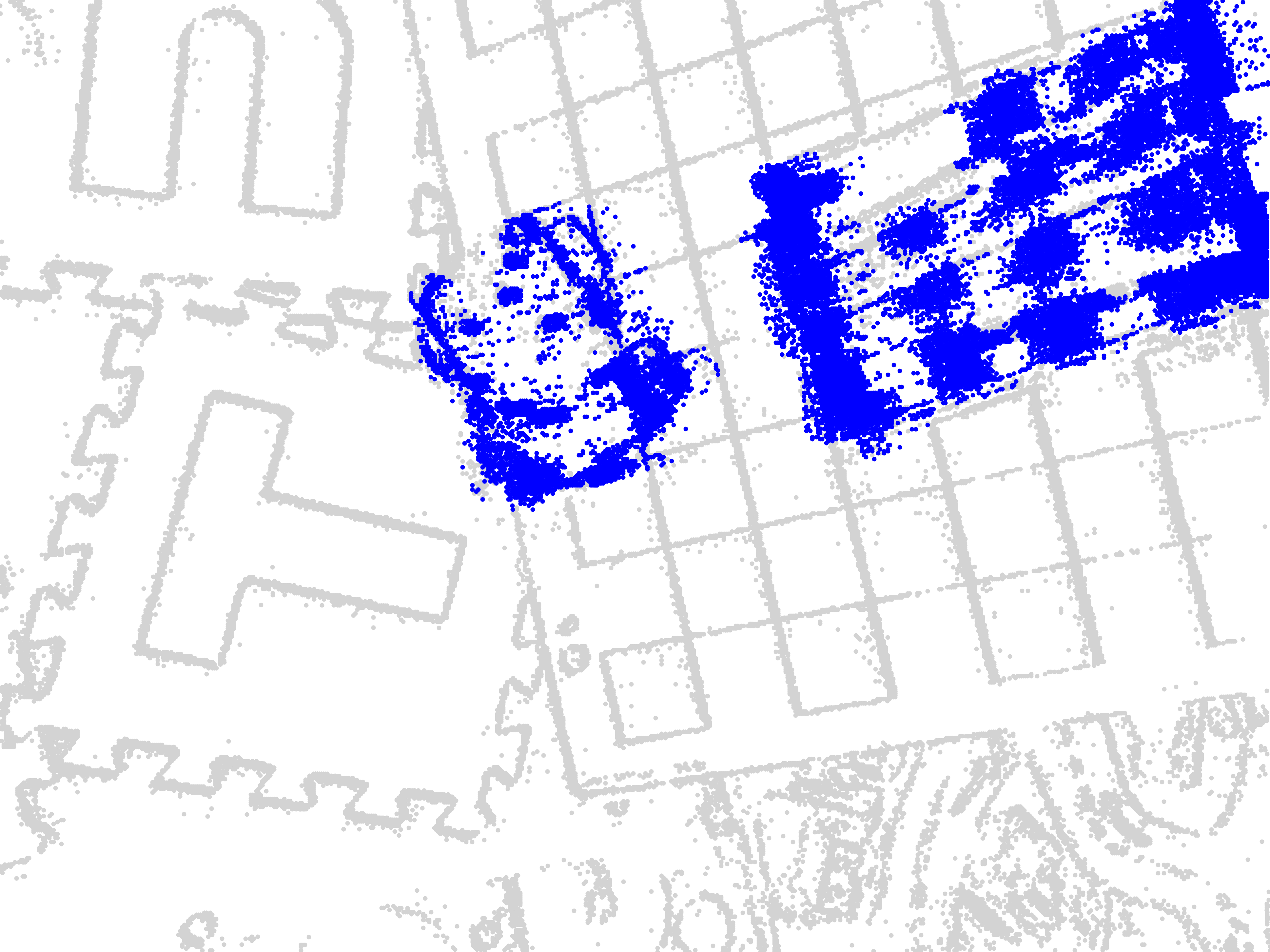} &
\includegraphics[width=0.2\linewidth]{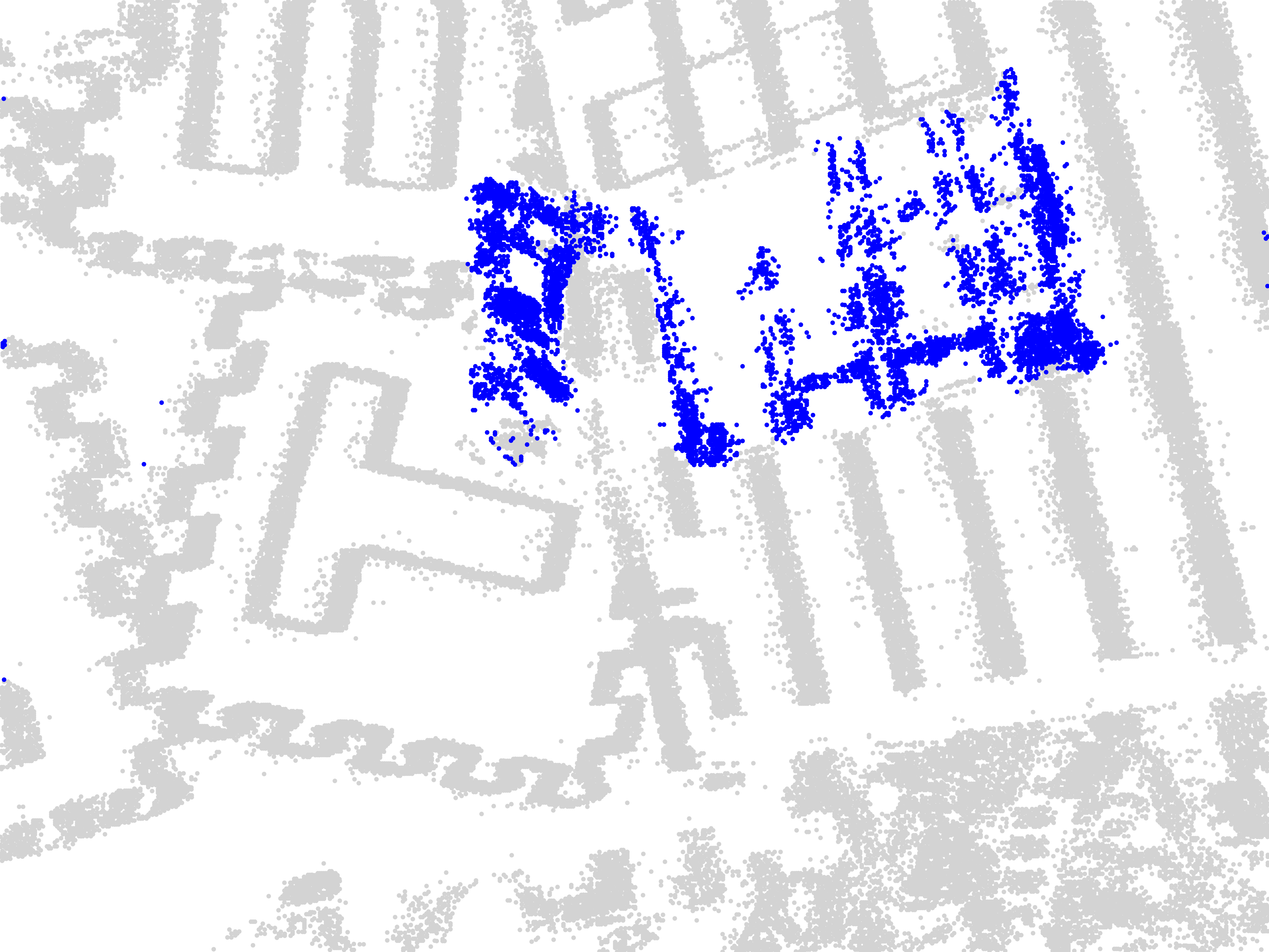} \\
 0.732 &  0.743 &  0.755 & 0.704  & 0.805 \\
\end{tabular}
\caption*{Frames from sequence 13-05}
\vspace{2mm}
\begin{tabular}{ccccc}
\includegraphics[width=0.2\linewidth]{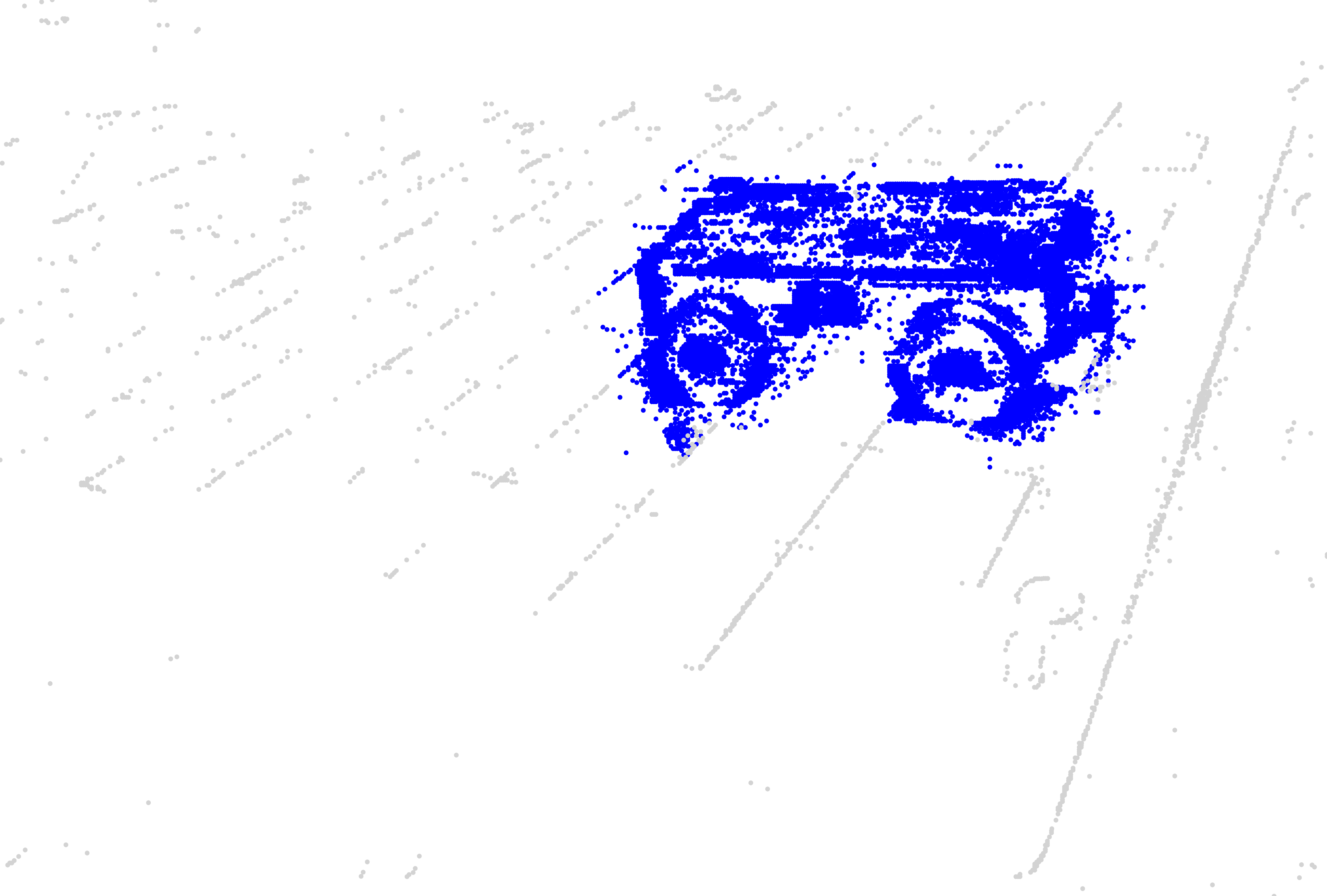} &
\includegraphics[width=0.2\linewidth]{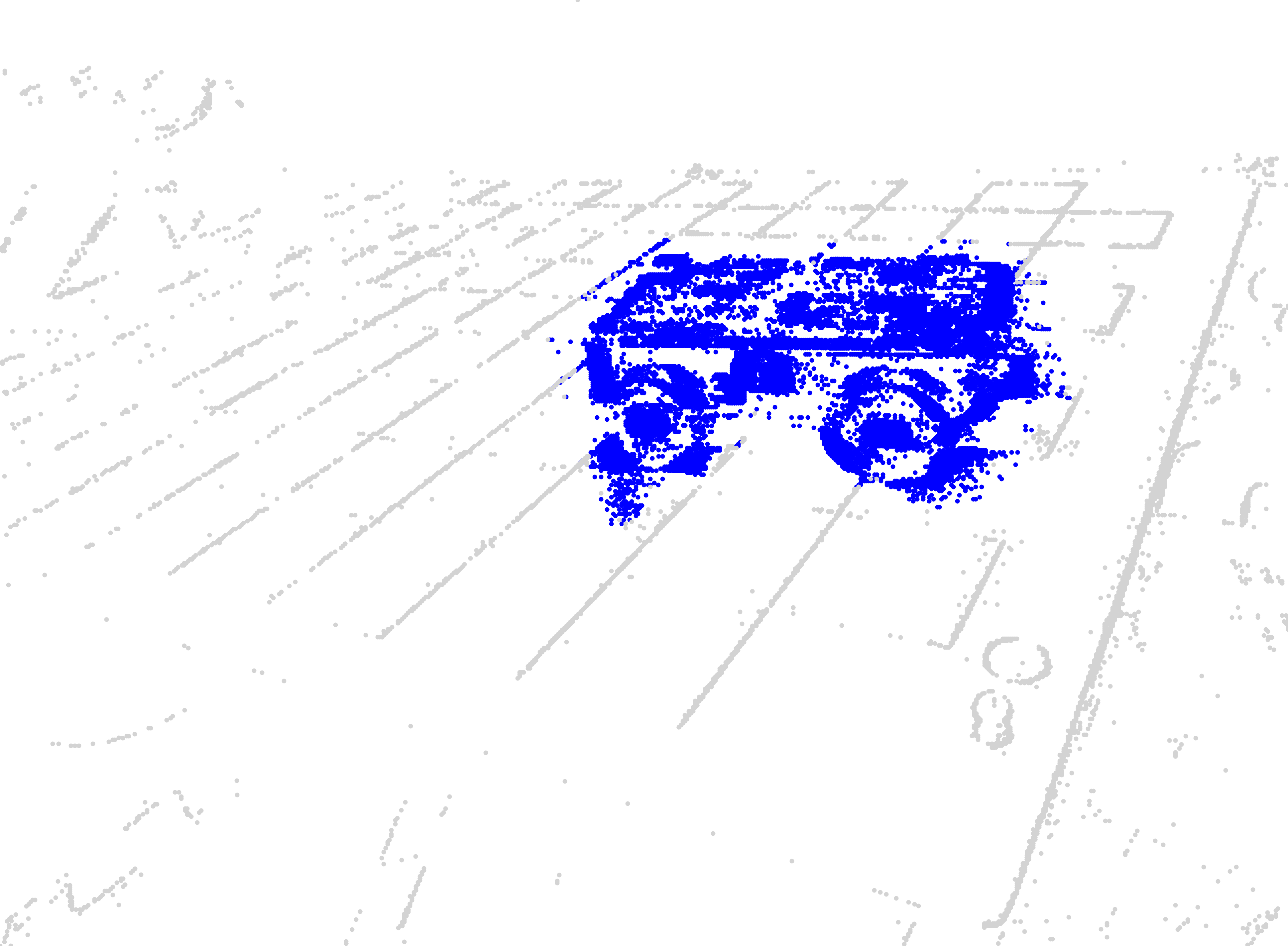} &
\includegraphics[width=0.2\linewidth]{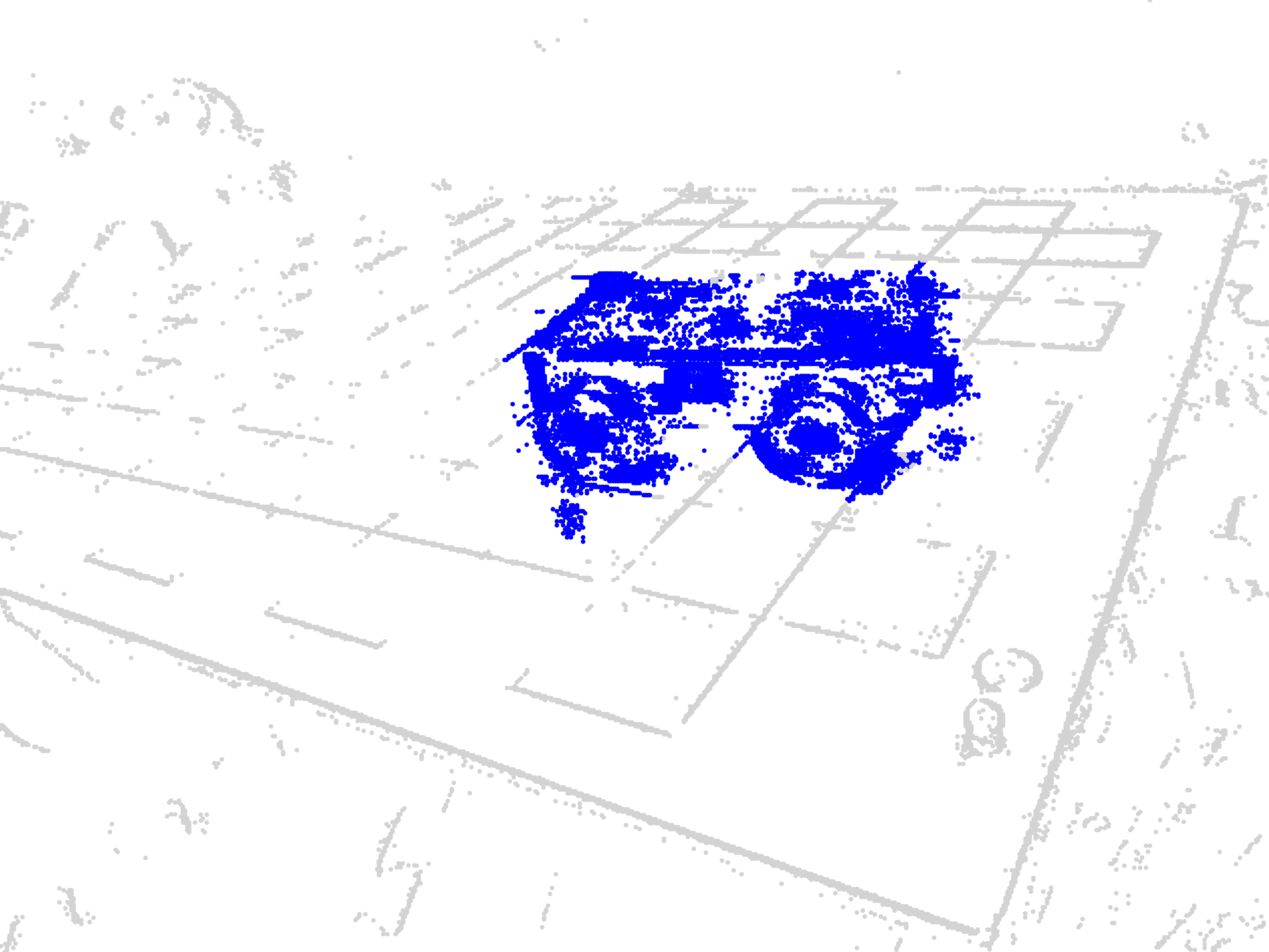} &
\includegraphics[width=0.2\linewidth]{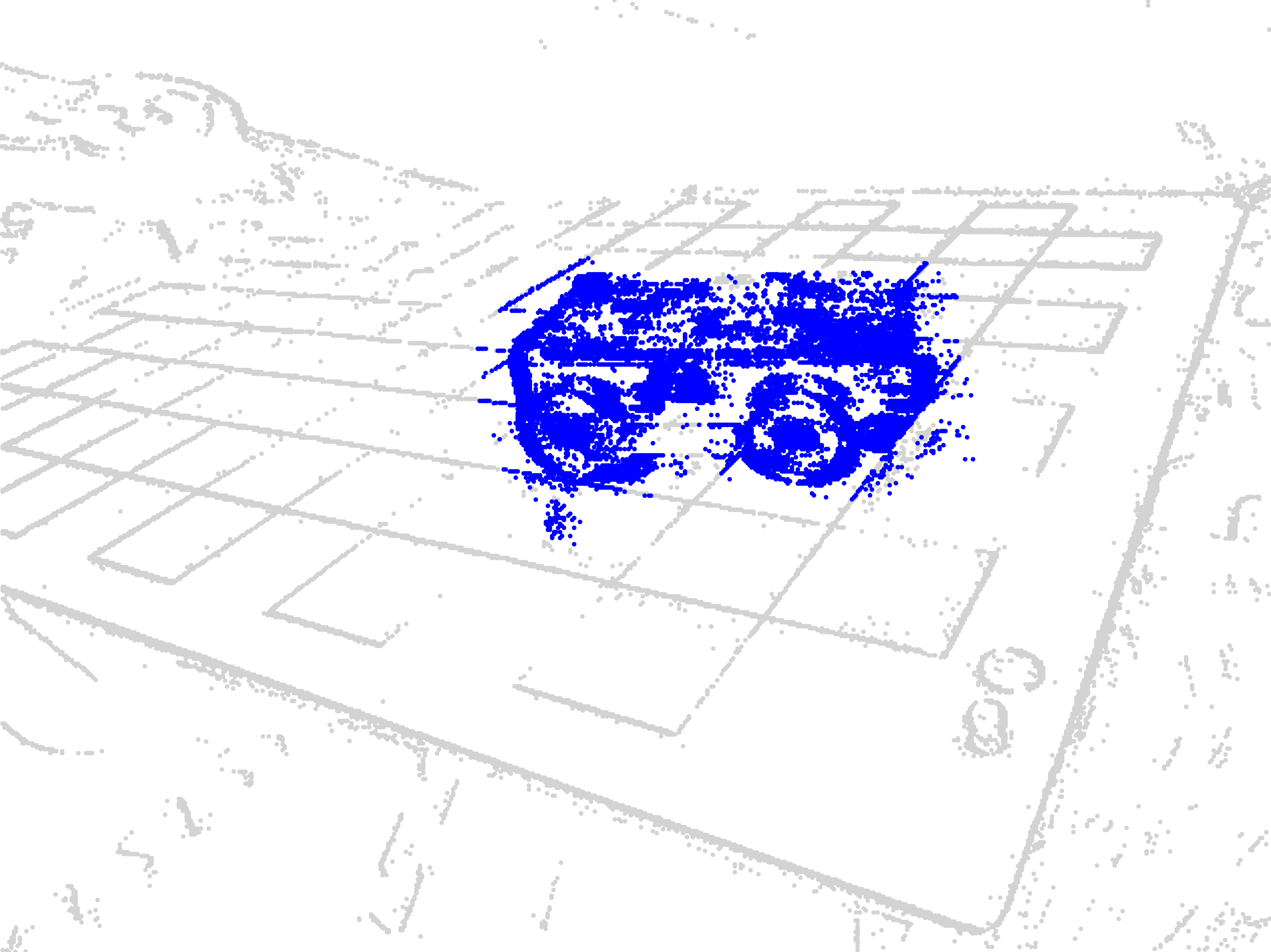} &
\includegraphics[width=0.2\linewidth]{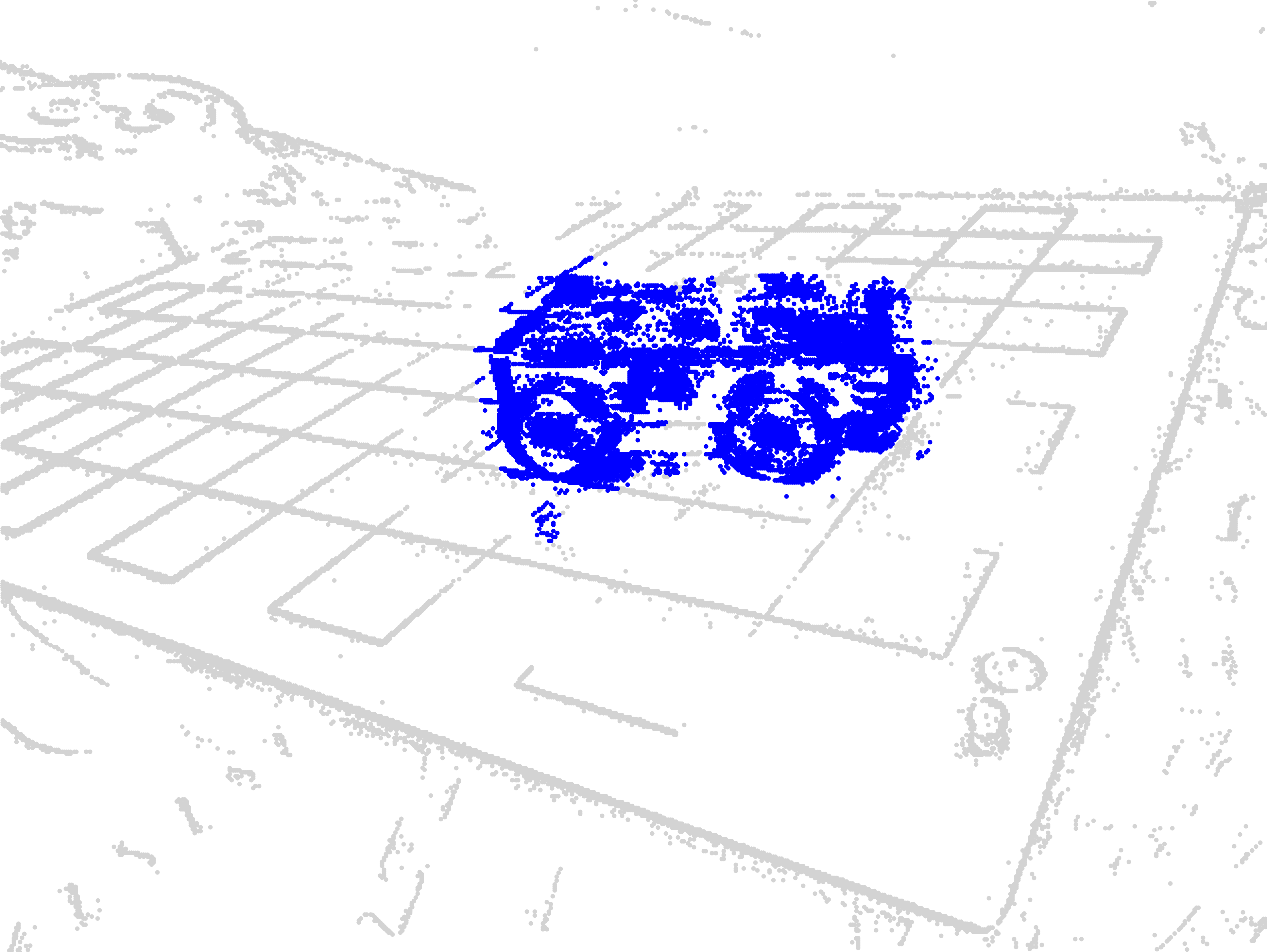} \\
 0.795 &  0.805 &  0.804 & 0.796  & 0.777 \\
\end{tabular}
\caption*{Frames from sequence 14-03}
\vspace{2mm}
\begin{tabular}{ccccc}
\includegraphics[width=0.2\linewidth]{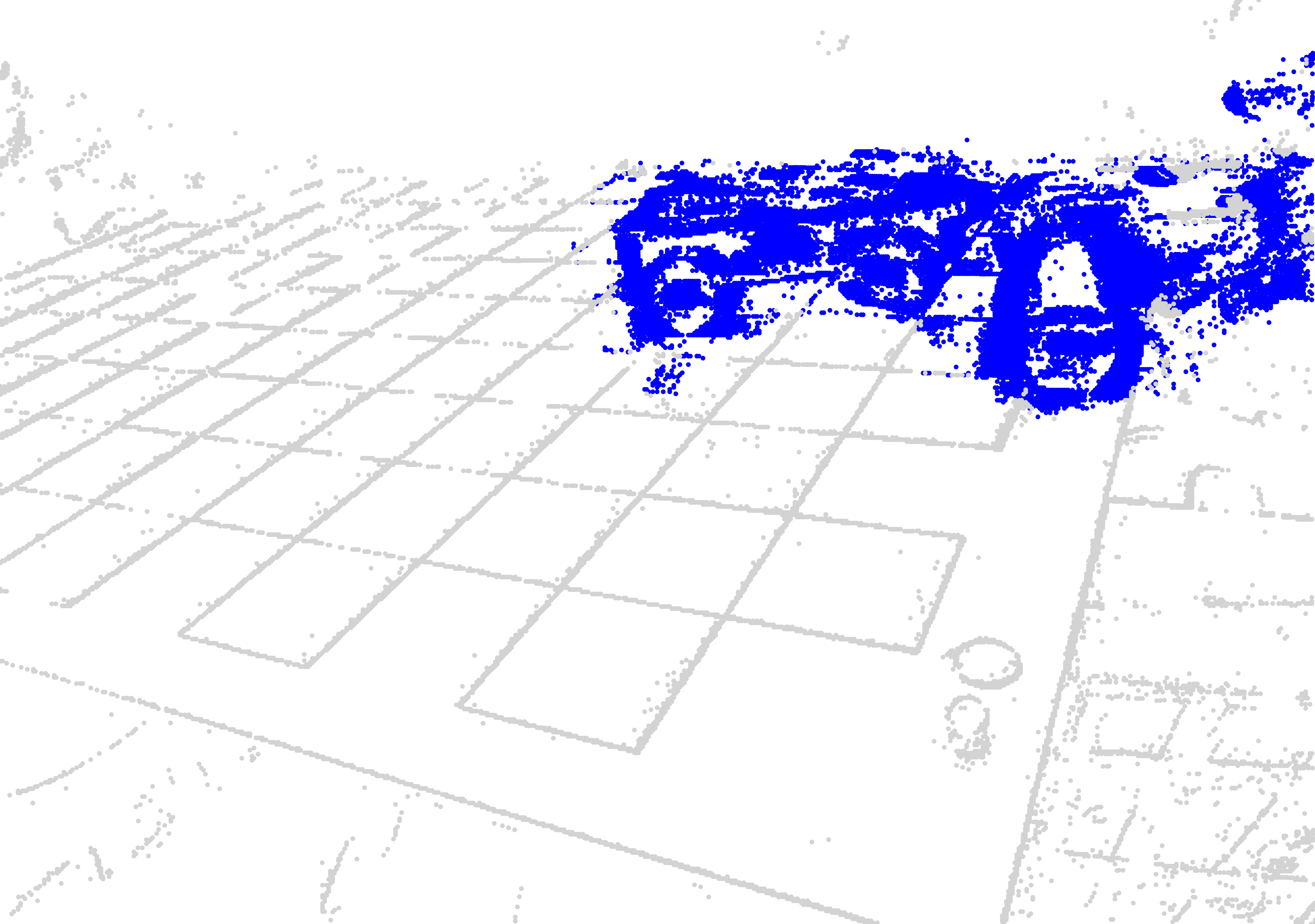} &
\includegraphics[width=0.2\linewidth]{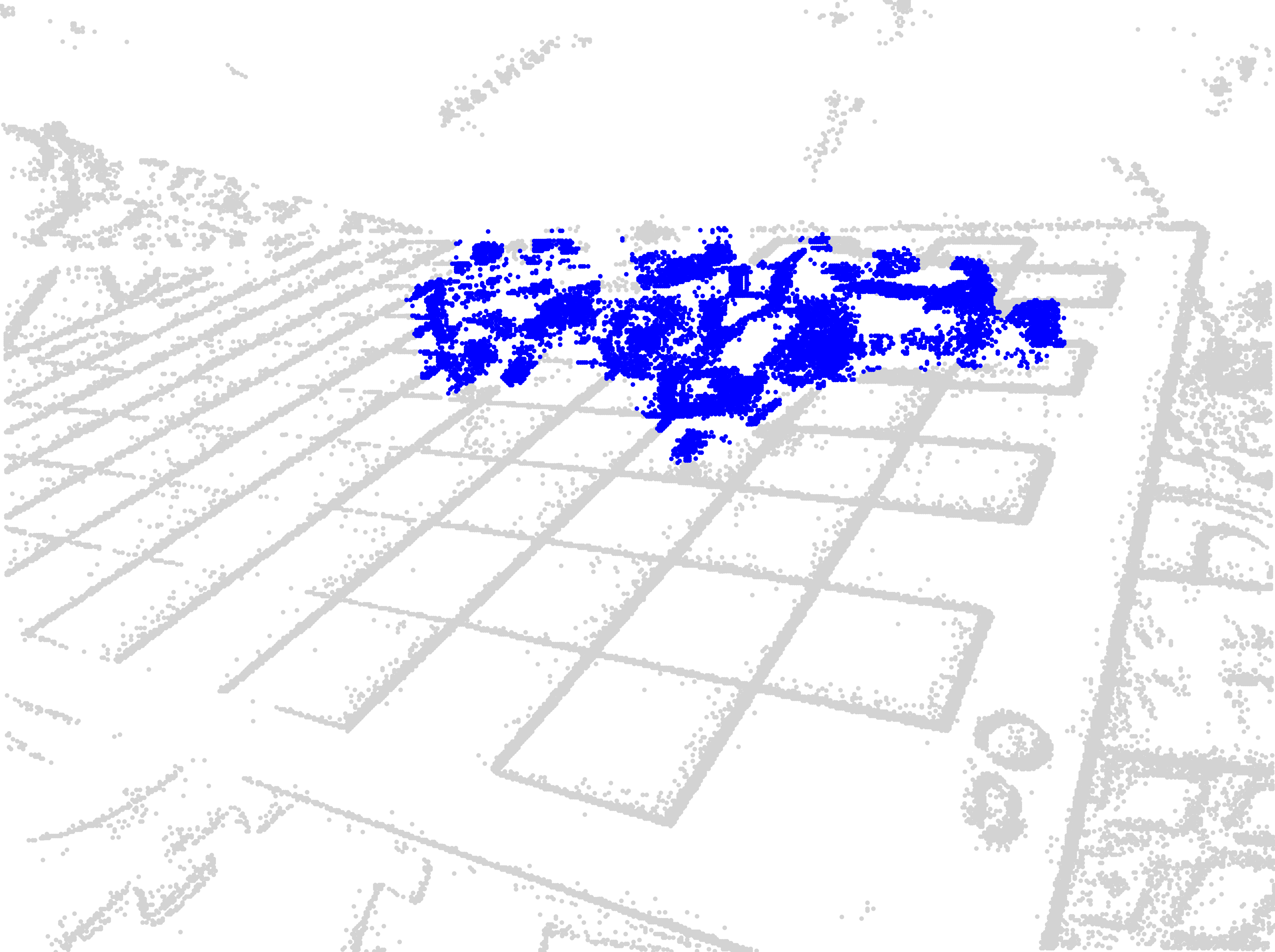} &
\includegraphics[width=0.2\linewidth]{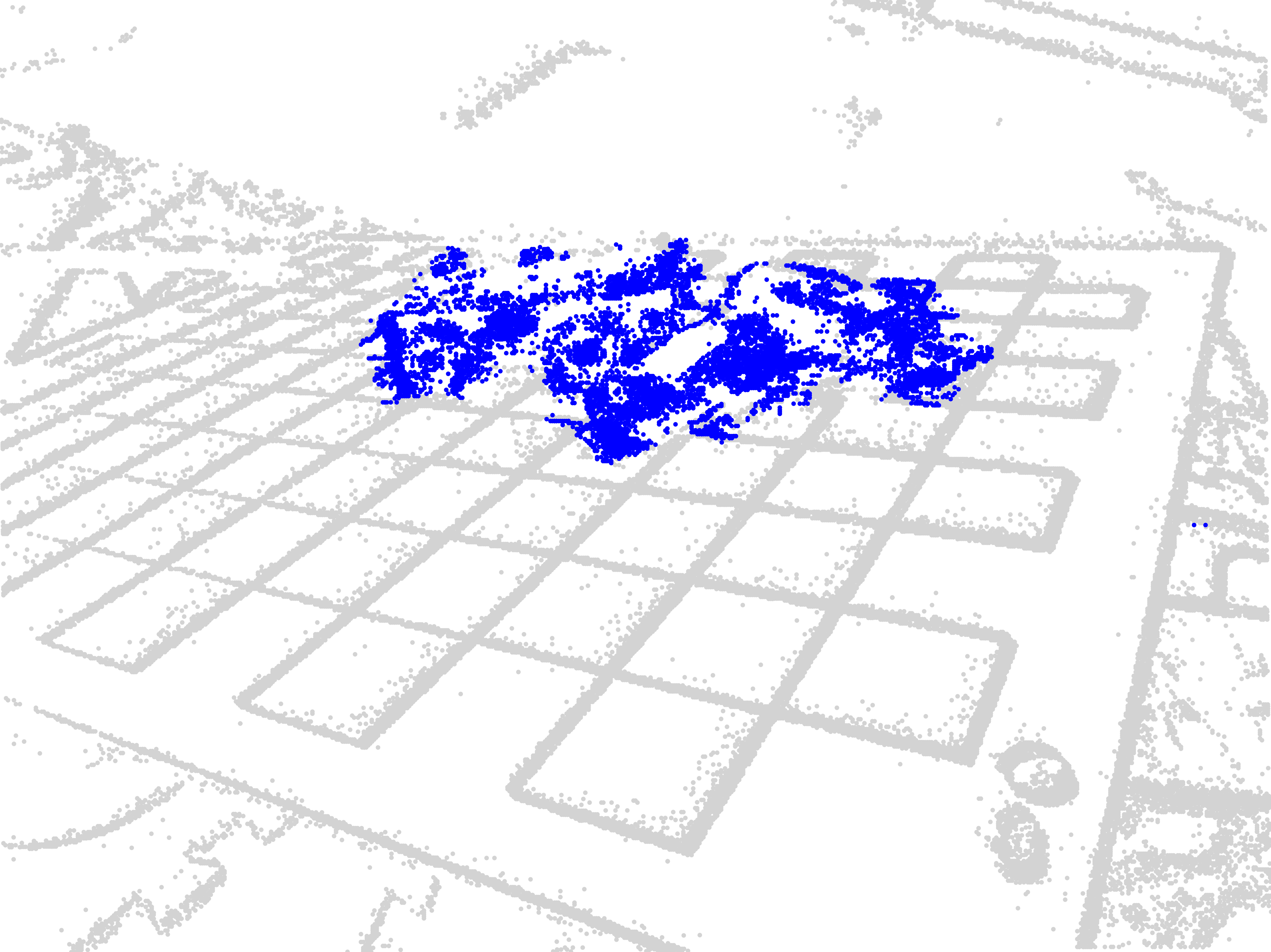} &
\includegraphics[width=0.2\linewidth]{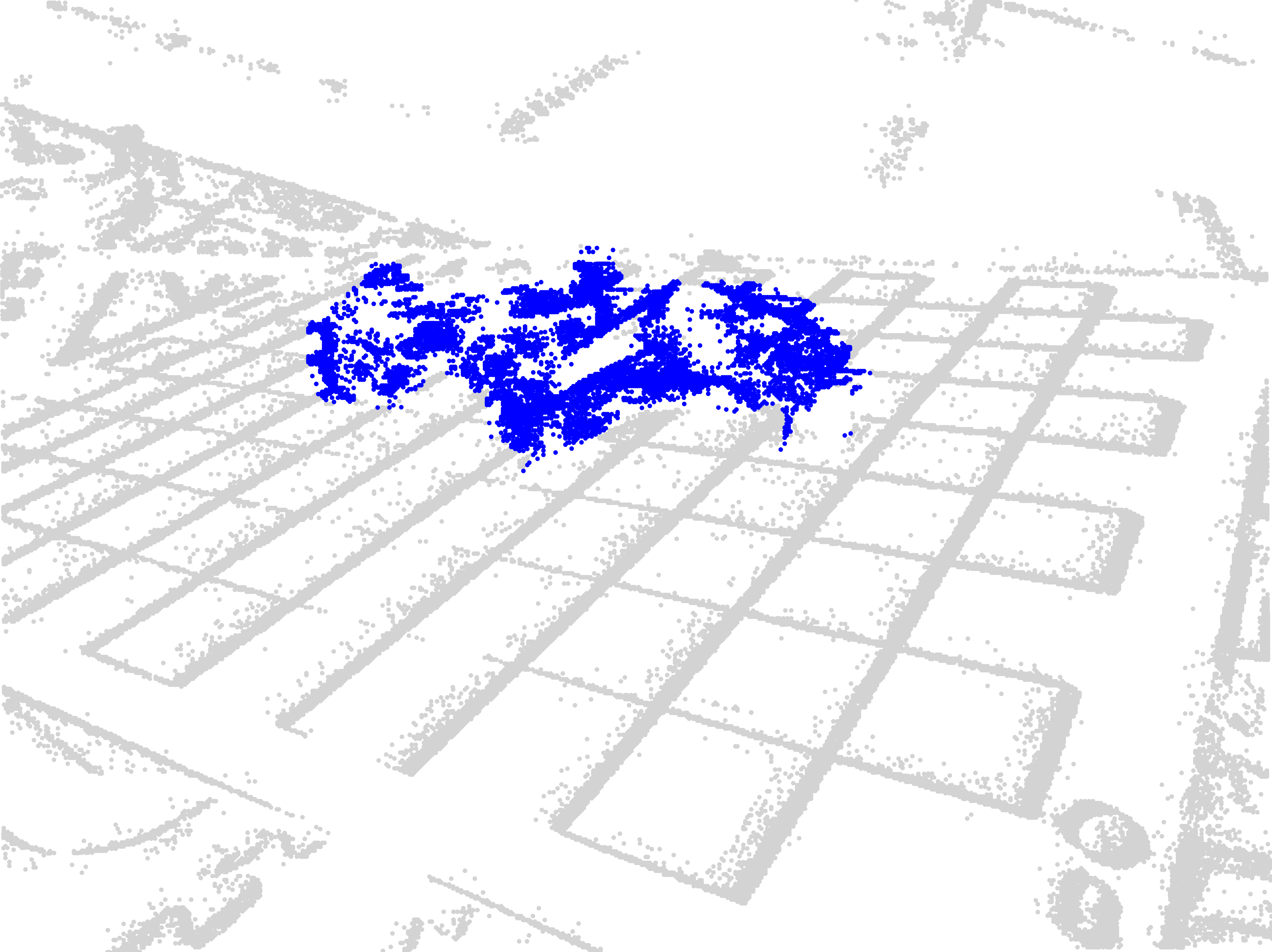} &
\includegraphics[width=0.2\linewidth]{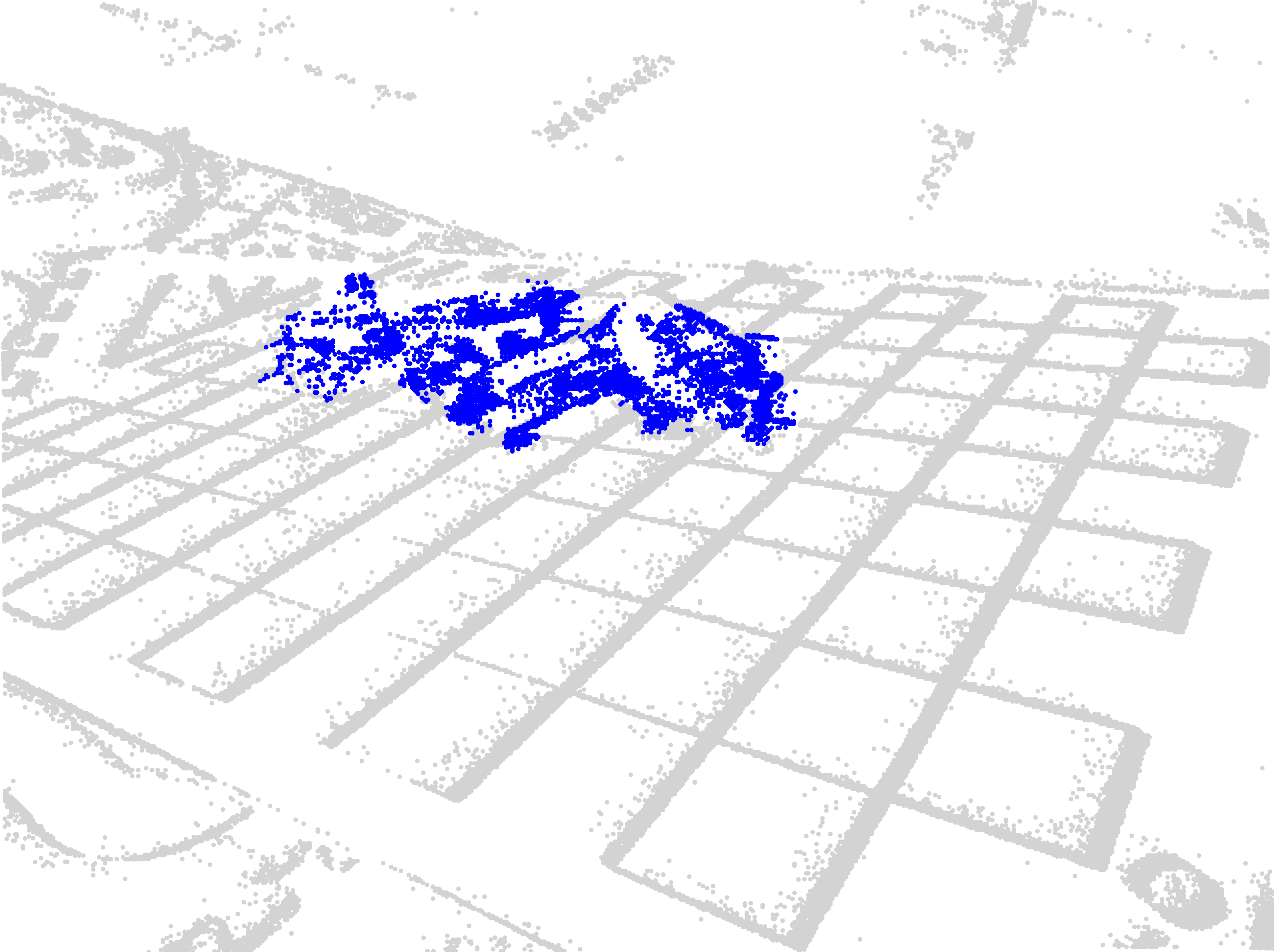} \\
 0.738 &  0.761 &  0.726 & 0.740  & 0.703 \\
\end{tabular}
\caption*{Frames from sequence 14-04}
\vspace{2mm}
\begin{tabular}{ccccc}
\includegraphics[width=0.2\linewidth]{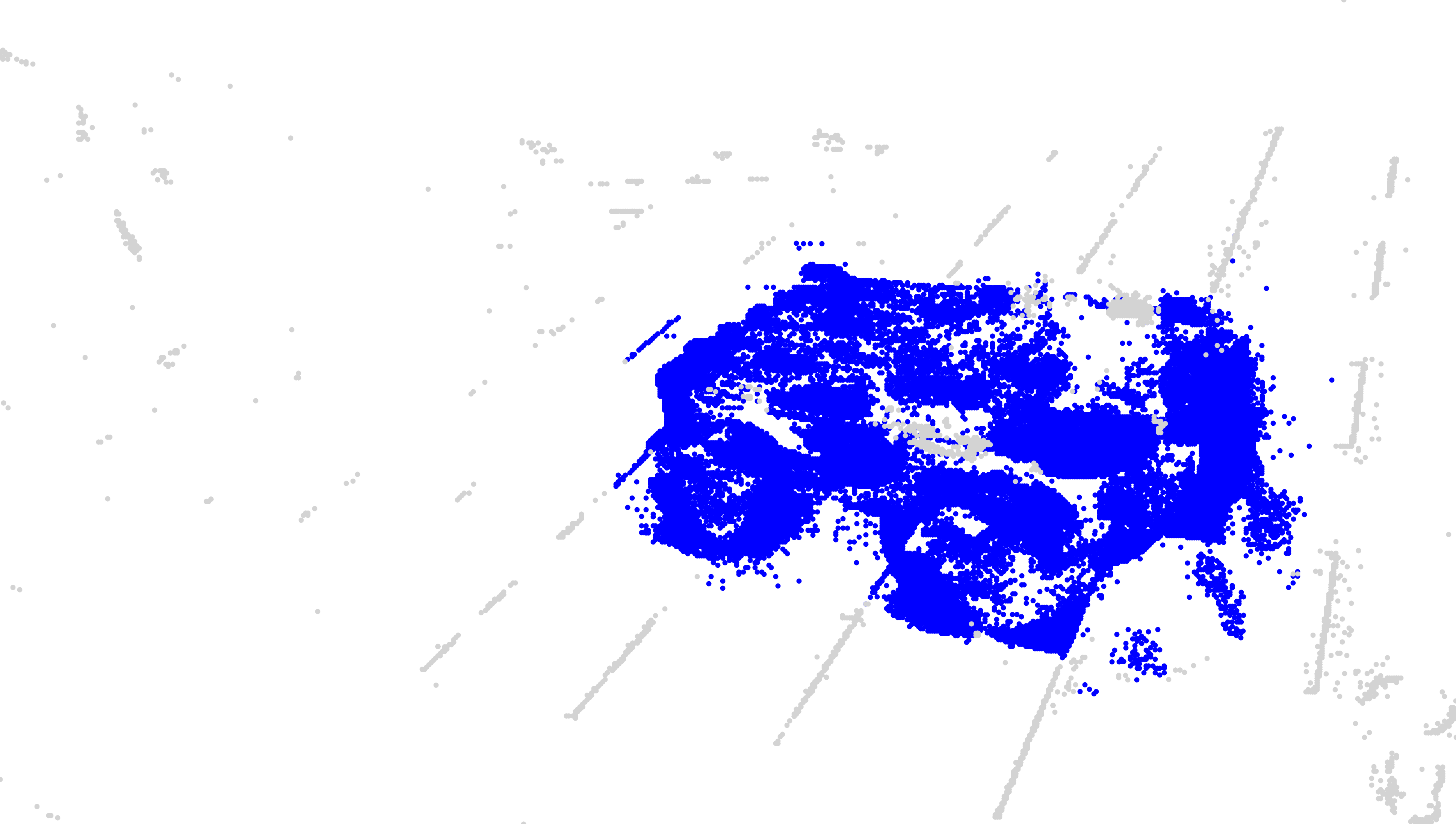} &
\includegraphics[width=0.2\linewidth]{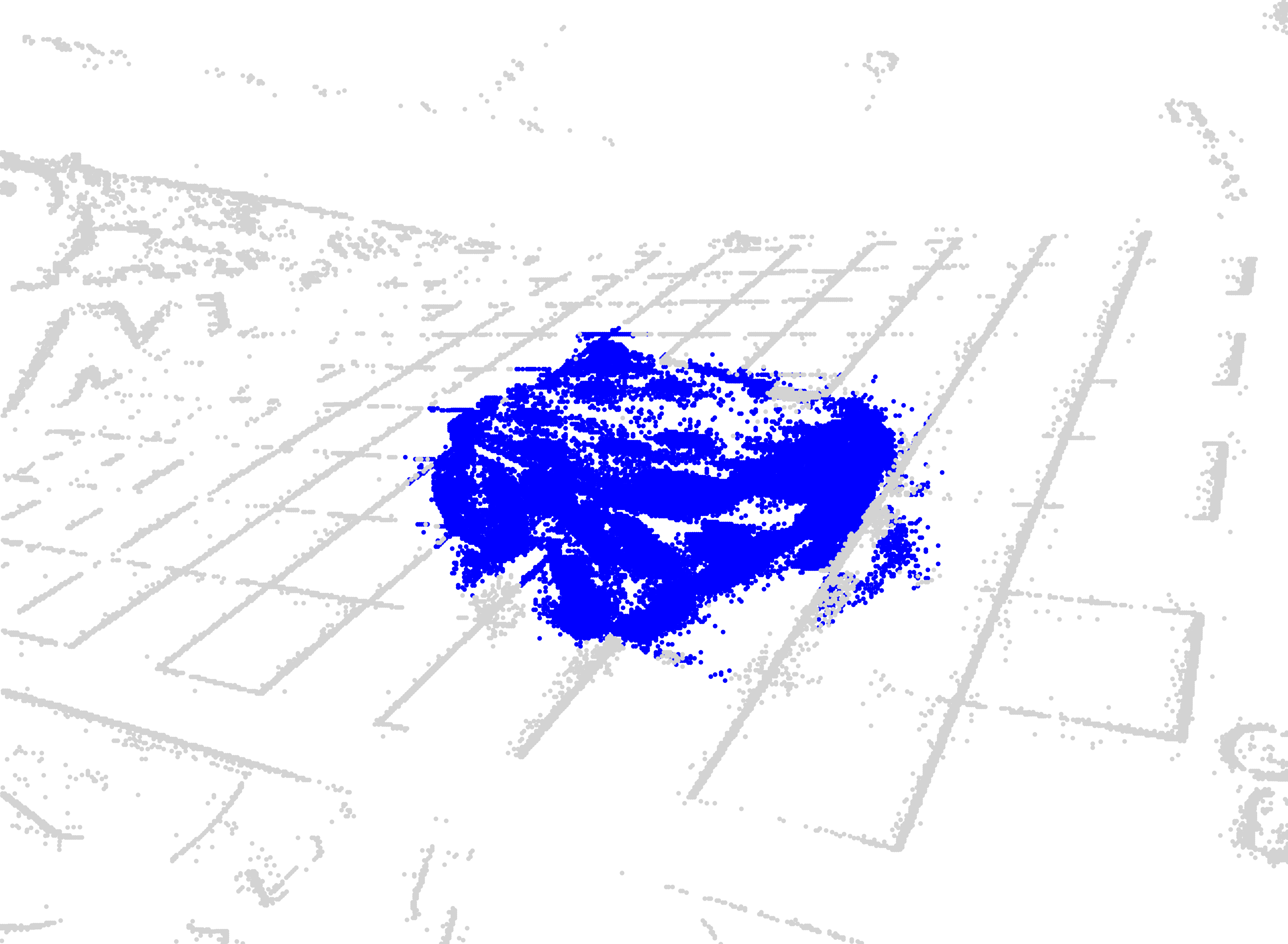} &
\includegraphics[width=0.2\linewidth]{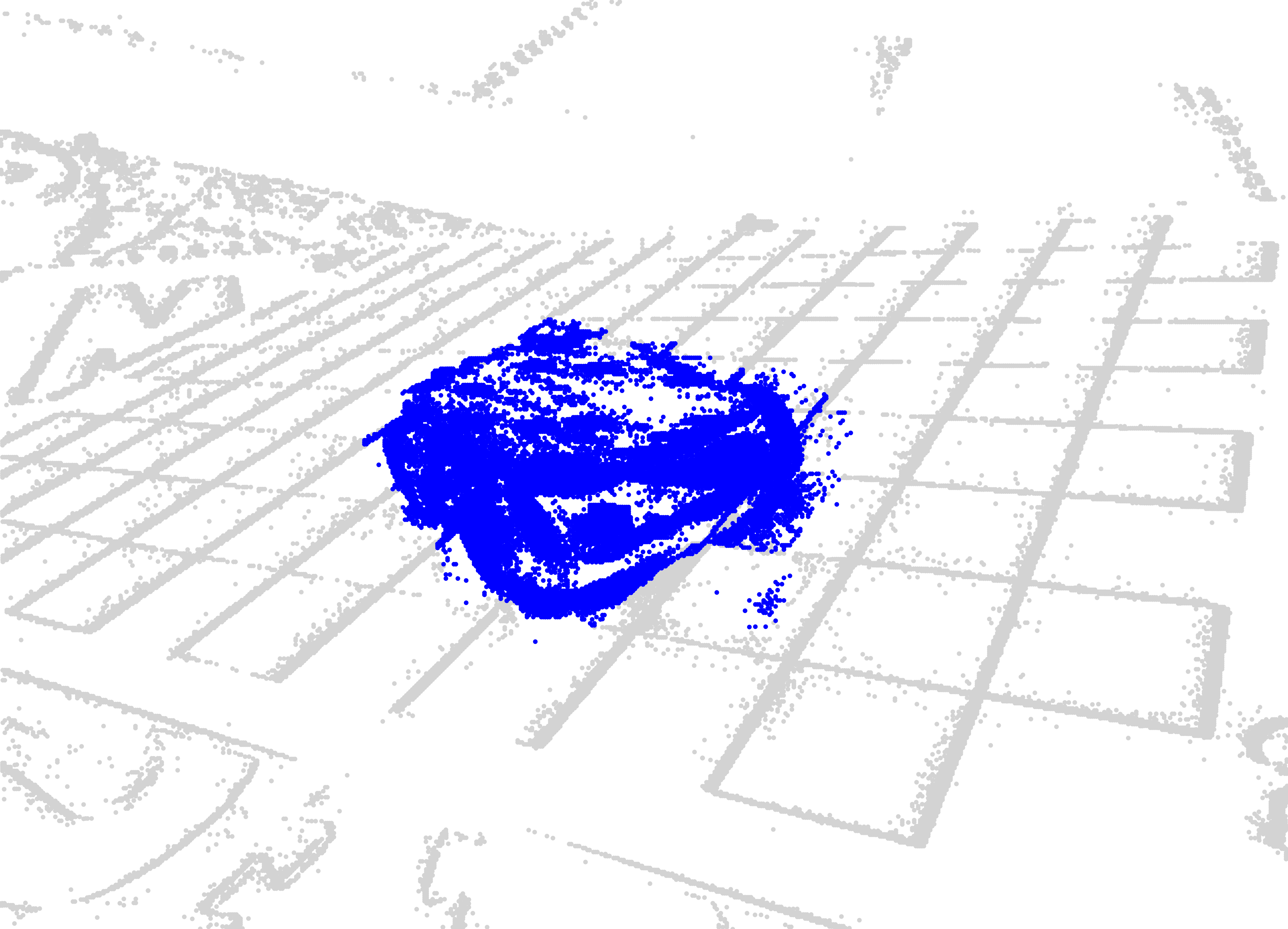} &
\includegraphics[width=0.2\linewidth]{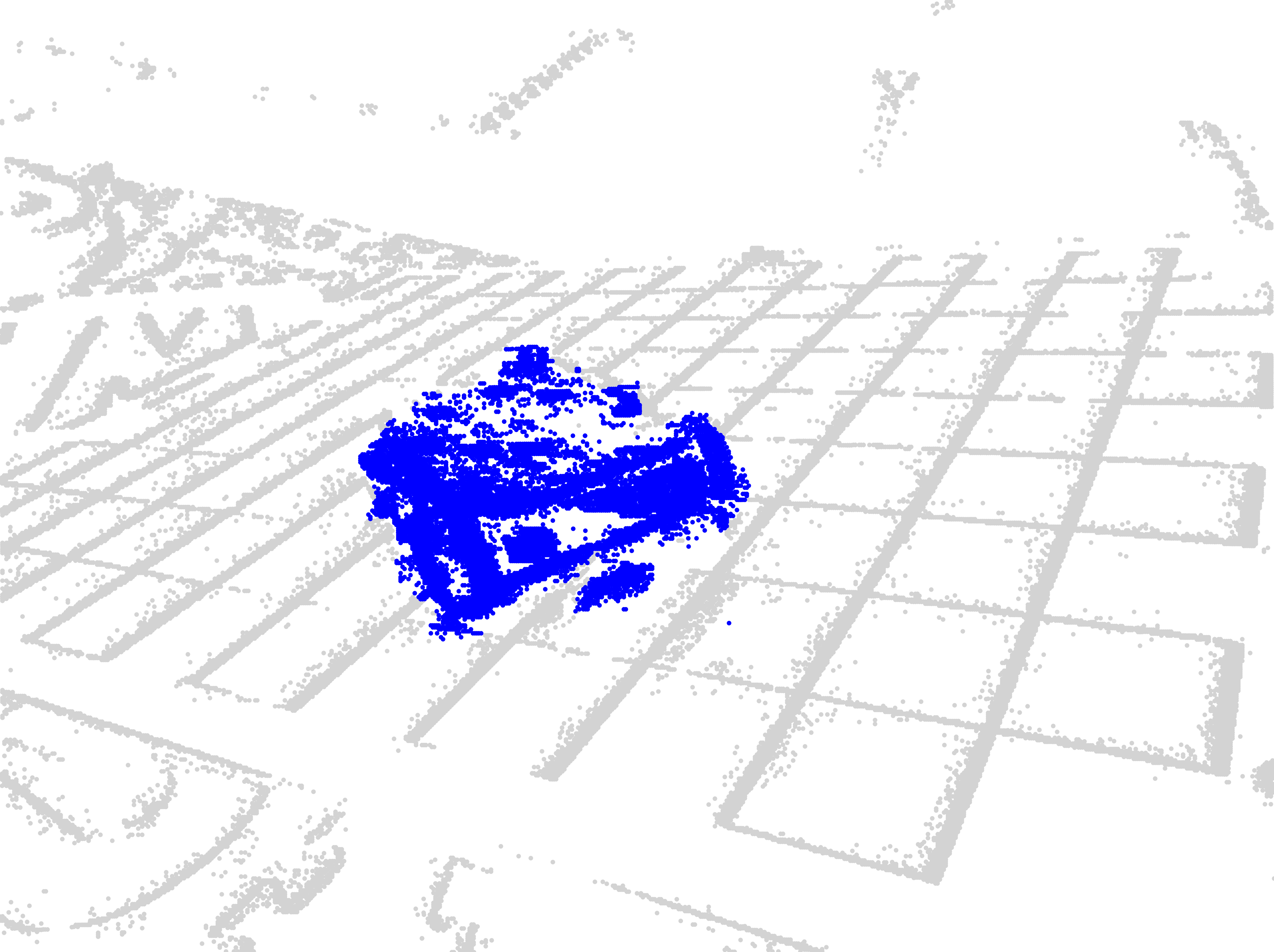} &
\includegraphics[width=0.2\linewidth]{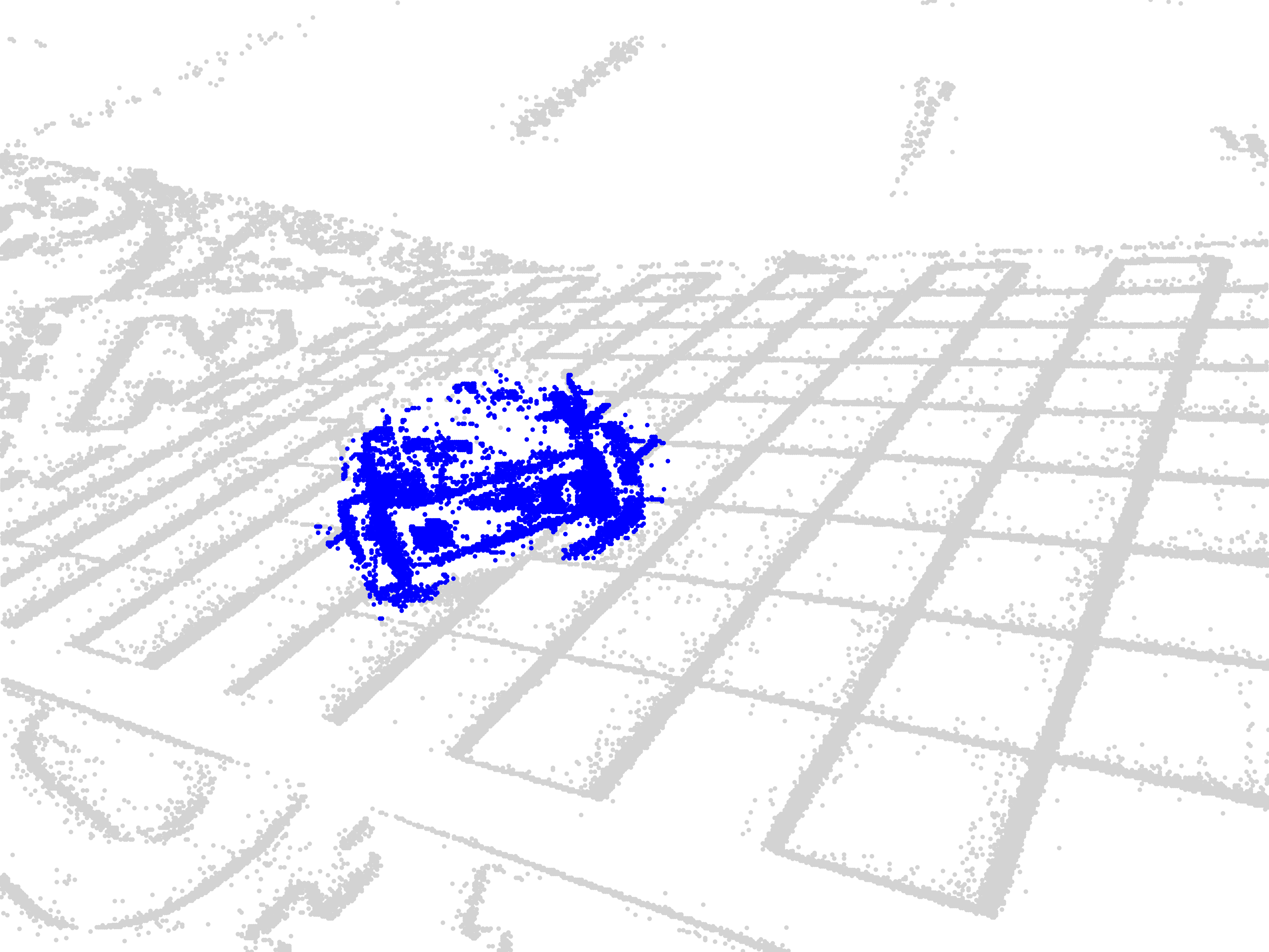} \\
 0.786 &  0.701 &  0.788 & 0.867  & 0.828 \\
\end{tabular}
\caption*{Frames from sequence 14-05}
\vspace{2mm}
\begin{tabular}{ccccc}
\includegraphics[width=0.2\linewidth]{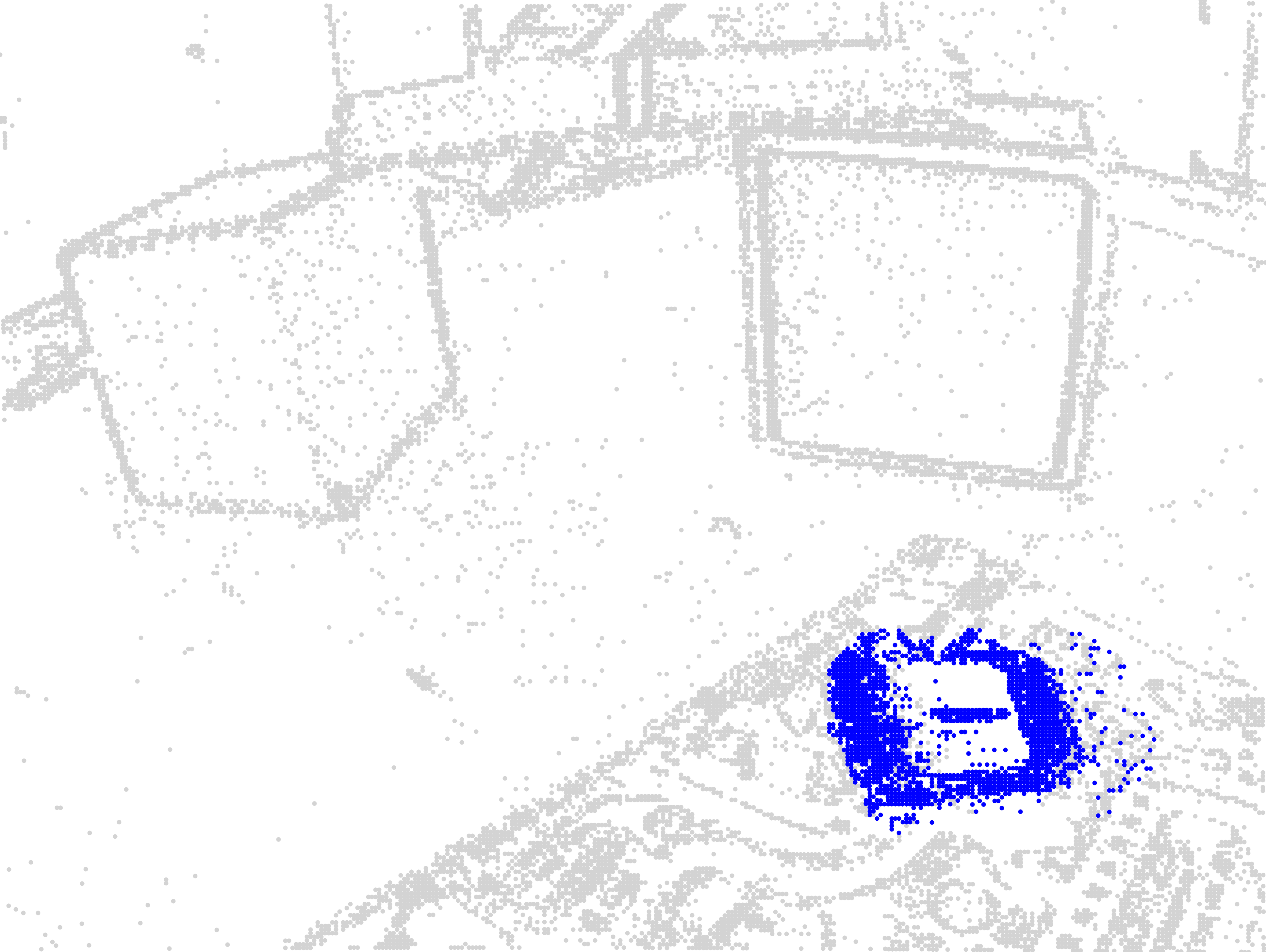} &
\includegraphics[width=0.2\linewidth]{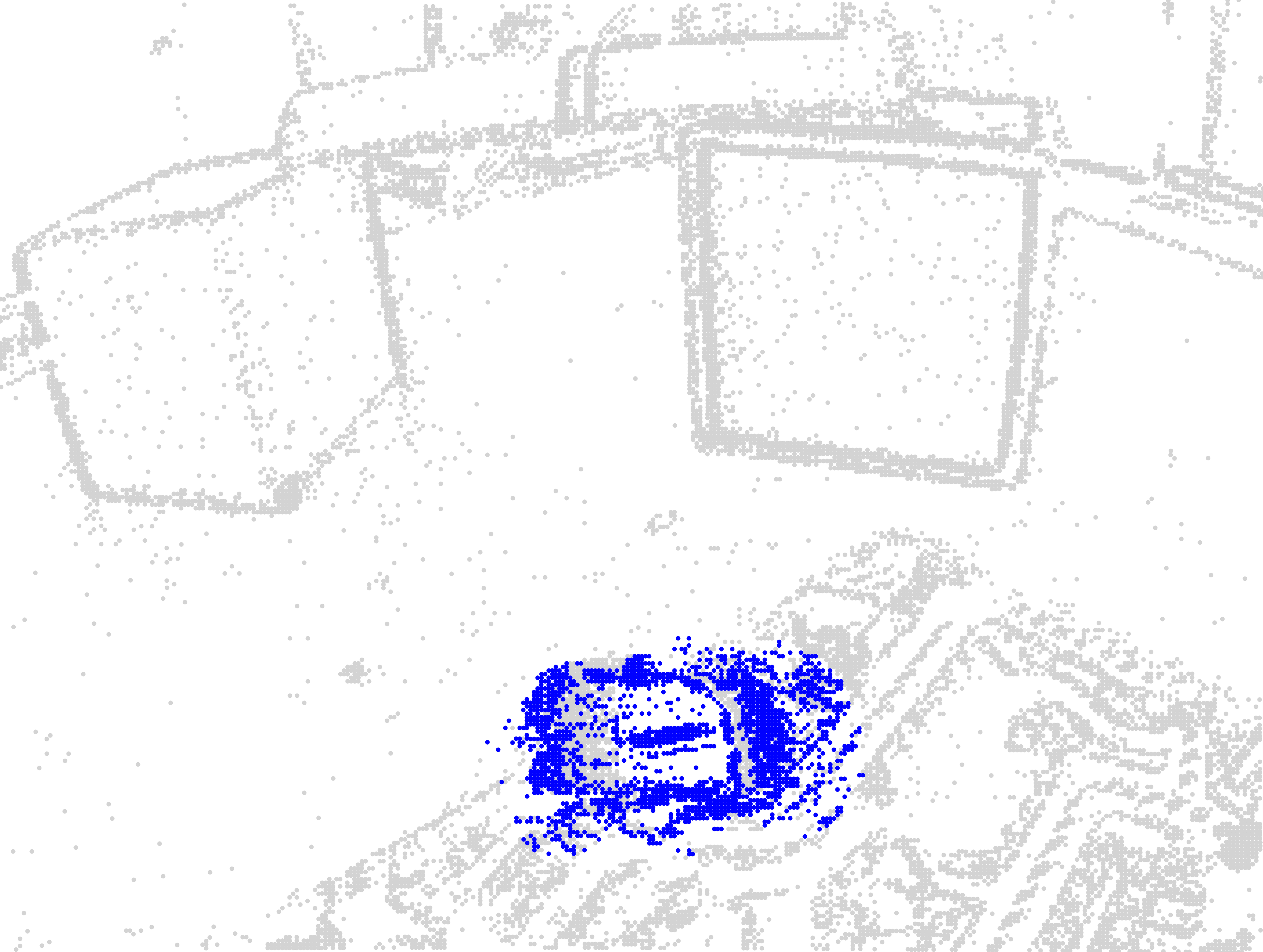} &
\includegraphics[width=0.2\linewidth]{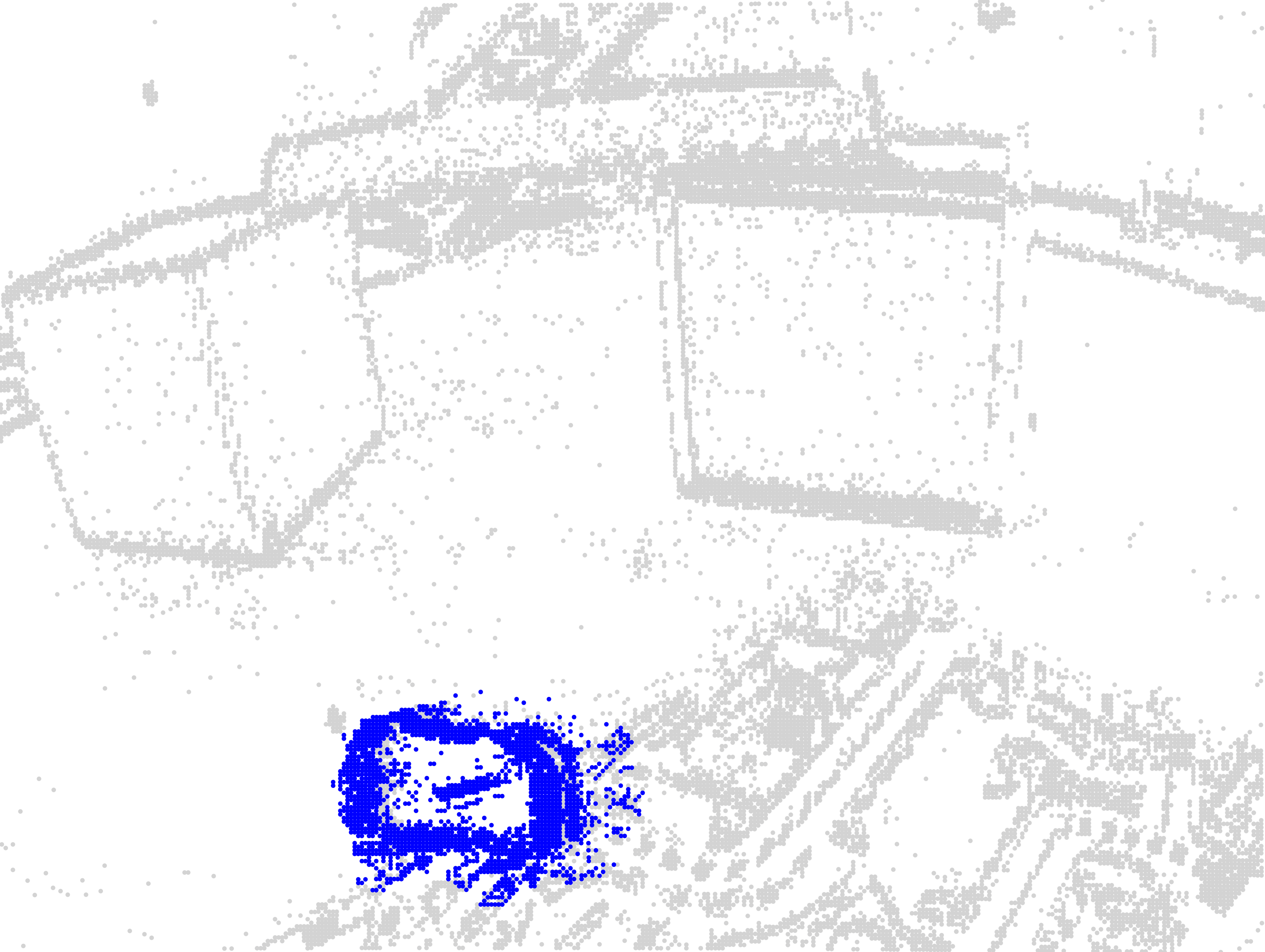} &
\includegraphics[width=0.2\linewidth]{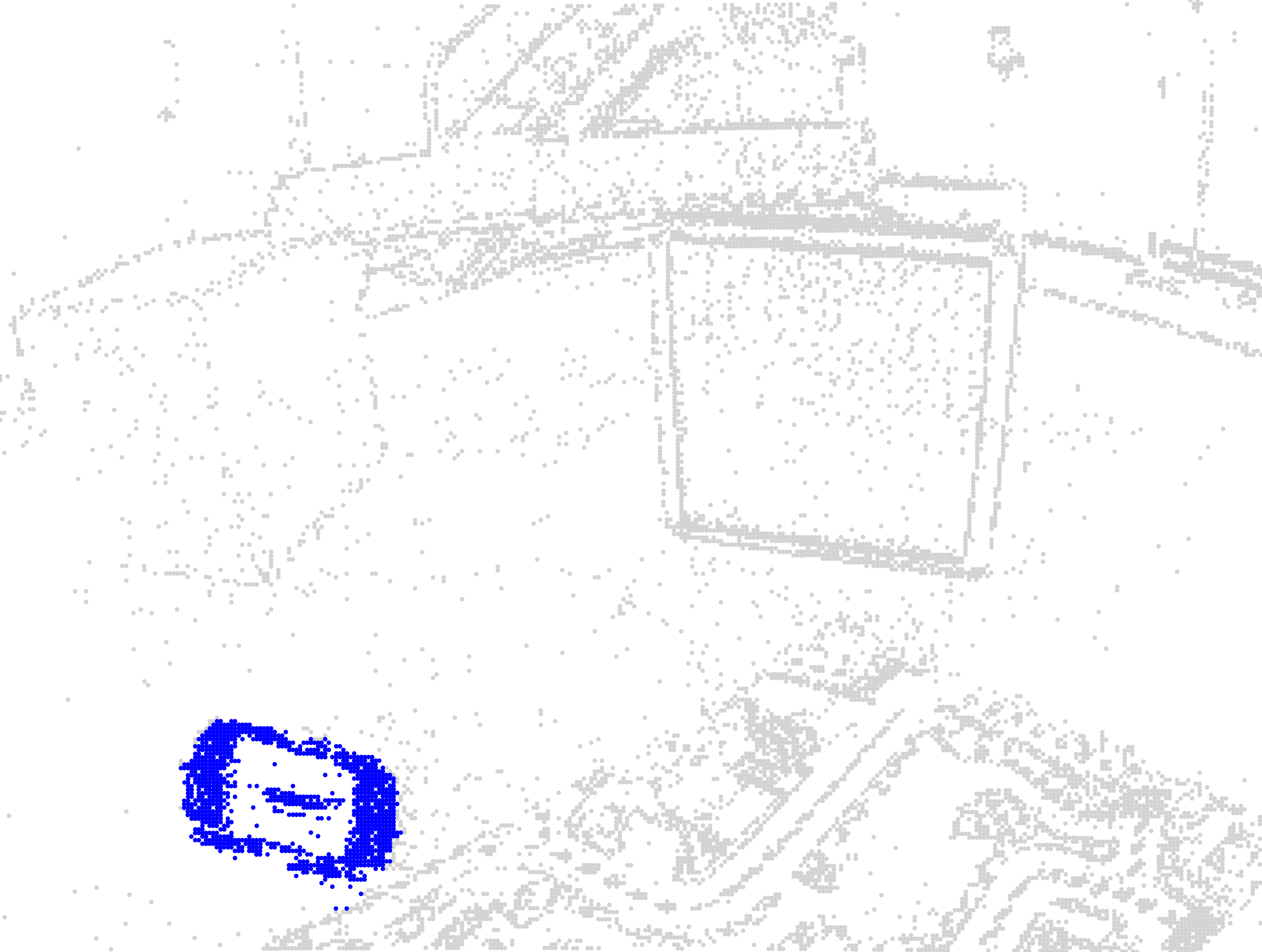} &
\includegraphics[width=0.2\linewidth]{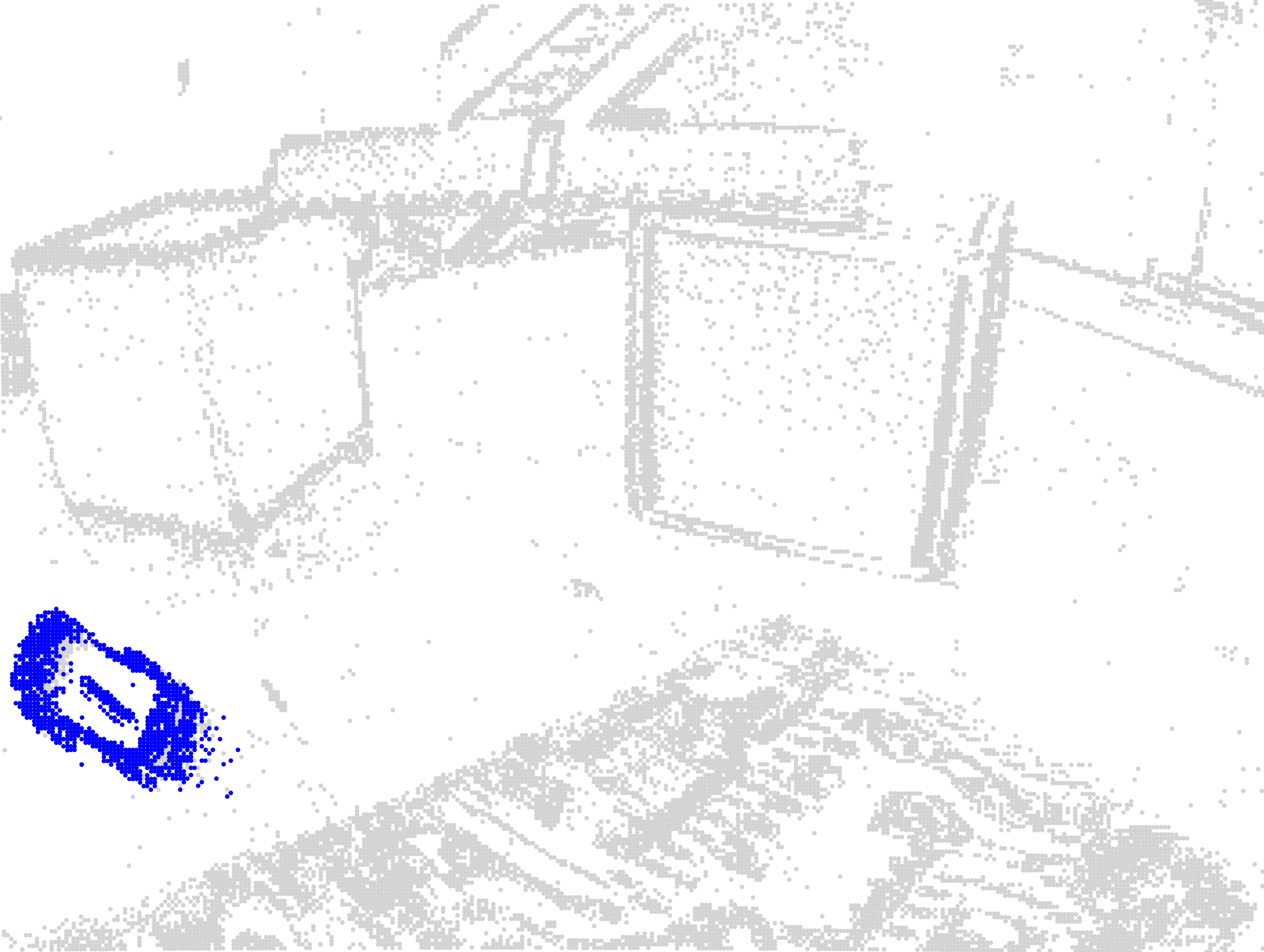} \\
 0.694 &  0.576 &  0.733 & 0.837  & 0.739 \\
\end{tabular}
\caption*{Frames from sequence box-00}
\vspace{2mm}
\caption{Qualitative segmentation results with per-frame IoU values across different sequences in EVIMO2v2 and EVIMO1}
\label{fig:qualitative_iou_all}
\end{figure}
\begin{table}[htbp]
  \centering
  \begin{tabular}{l r}
    \toprule
    \textbf{Scene} & \textbf{IoU (\%)} \\
    \midrule
    13-00 & 82.15 \\
    13-05 & 76.24 \\
    14-03 & 79.58 \\
    14-04 & 73.36 \\
    14-05 & 75.67 \\
    \bottomrule
  \end{tabular}
  \caption{IoU Evaluation Across Different Scenes}
  \label{tab:iou_evaluation}
\end{table}

Table \ref{tab:iou_evaluation} presents the Intersection over Union (IoU) evaluation of our proposed method across five different scenes from the  EVIMO2v2 \cite{burner2022evimo2} dataset. IoU is computed as the ratio of the intersection to the union between the predicted and ground truth events, evaluated only in frames containing moving object(s). 

We also provide a qualitative evaluation of the segmentation performance in Figure \ref{fig:qualitative_iou_all}.

\subsection{Translation Accuracy Evaluation}
\label{sec:translation_evaluation}

To assess the accuracy of our motion estimation approach, we compare the predicted per-frame translational motion of the camera egomotion, and of the objects against the ground truth motion obtained from the full SE(3) camera and objects pose data. Specifically, we evaluate the translational accuracy of the camera egomotion, and the accuracy of the estimated 2D image-plane motion induced by the 3D translation, focusing on the horizontal ($\Delta X$) and vertical ($\Delta Y$) components separately.

\paragraph{Camera Egomotion Evaluation}

We quantitatively evaluate the accuracy of our approach's translational velocity estimation, as the IMU directly measures rotational motion. Table~\ref{tab:egomotion} reports the per-axis Root Mean Square Error (RMSE) of translational velocity across three distinct scenes from the EVIMO2v2 dataset. In these scenes, an IMO is present, and our method estimates the camera's translational velocity.

\begin{table}[h]
  \centering
  \begin{tabular}{lccc}
    \toprule
    Scene & $V_x\!\downarrow$ (m/s) & $V_y\!\downarrow$ (m/s) & $V_z\!\downarrow$ (m/s) \\
    \midrule
    \texttt{13-05} & 0.08 & 0.05 & 0.02 \\
    \texttt{14-03} & 0.05 & 0.03 & 0.09 \\
    \texttt{14-04} & 0.06 & 0.03 & 0.05 \\
    \bottomrule
  \end{tabular}
  \caption{Per‐axis RMSE in velocity for scenes on EVIMO2v2.}
  \label{tab:egomotion}
\end{table}

These results confirm that our method reliably estimates the translational camera egomotion across different scenes in EVIMO2v2.

\paragraph{Object Translation Accuracy}

Given the ground truth object trajectory $\mathbf{T}_{wo}(t) \in SE(3)$ and the camera trajectory $\mathbf{T}_{wc}(t) \in SE(3)$, we compute the relative object pose in the camera frame as $\mathbf{T}_{co}(t)$.
From the relative motion between consecutive frames $\Delta \mathbf{T}_{co}(t) = \mathbf{T}_{co}(t+1) \cdot \mathbf{T}_{co}^{-1}(t)$, we extract the twist vector $\xi(t) = [\mathbf{v}(t)^\top, \boldsymbol{\omega}(t)^\top]^\top$ via the logarithmic map. 

We then compute the image-plane motion induced by 3D translation and rotation as:
\begin{equation}
    \dot{\mathbf{u}}(t) = A(x(t), y(t)) \cdot \mathbf{v}(t) + B(x(t), y(t)) \cdot \boldsymbol{\omega}(t)
\end{equation}
where $A(x, y)$ and $B(x, y)$ are projection matrices that encode translation and rotation effects, respectively. Our method estimates only the translational component $A(x, y)\cdot \mathbf{v}(t)$ based on normal flow and segment-wise motion consistency, and does not explicitly account for $\boldsymbol{\omega}(t)$. Since the estimation of motion along the depth axis ($\Delta Z$) is highly inaccurate, our analysis focuses only on the translation parallel to the image plane.

\begin{figure}[htbp]
    \centering

    \begin{subfigure}[t]{0.49\linewidth}
        \centering
        \includegraphics[width=\linewidth]{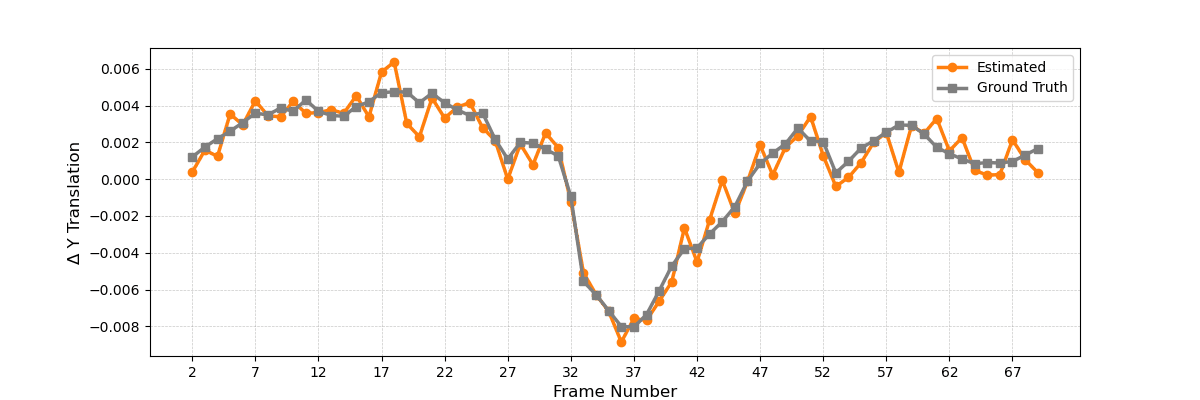}
        \caption*{13-05: $\Delta X$ Translation}
    \end{subfigure}
    \hfill
    \begin{subfigure}[t]{0.49\linewidth}
        \centering
        \includegraphics[width=\linewidth]{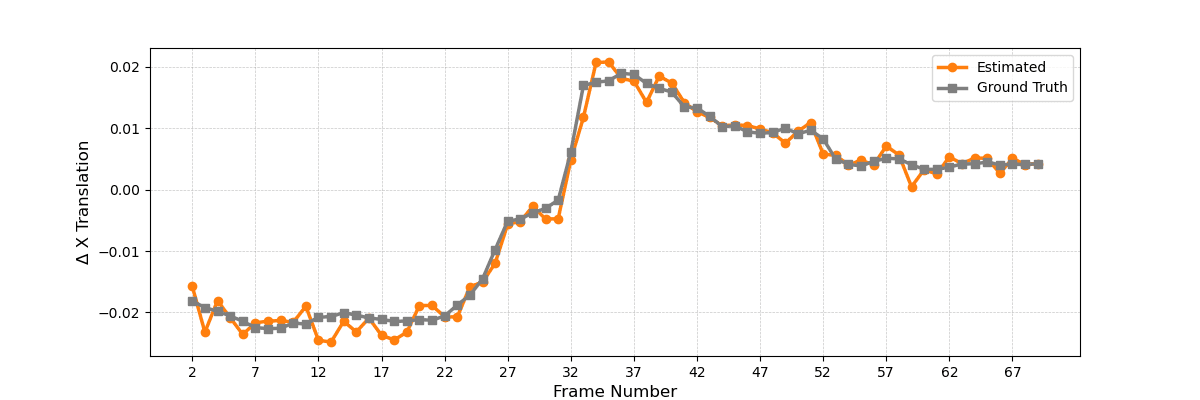}
        \caption*{13-05: $\Delta Y$ Translation}
    \end{subfigure}

    \vspace{1em} 

    \begin{subfigure}[t]{0.49\linewidth}
        \centering
        \includegraphics[width=\linewidth]{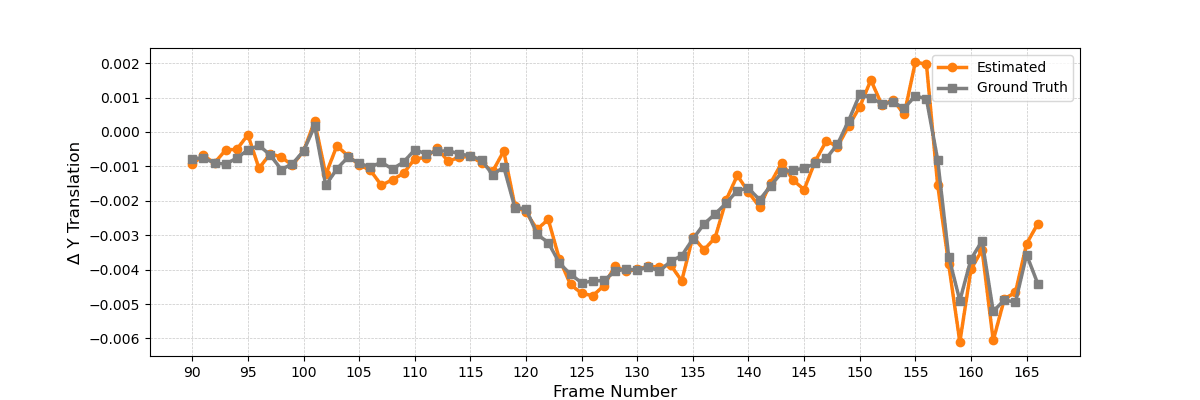}
        \caption*{14-03: $\Delta X$ Translation}
    \end{subfigure}
    \hfill
    \begin{subfigure}[t]{0.49\linewidth}
        \centering
        \includegraphics[width=\linewidth]{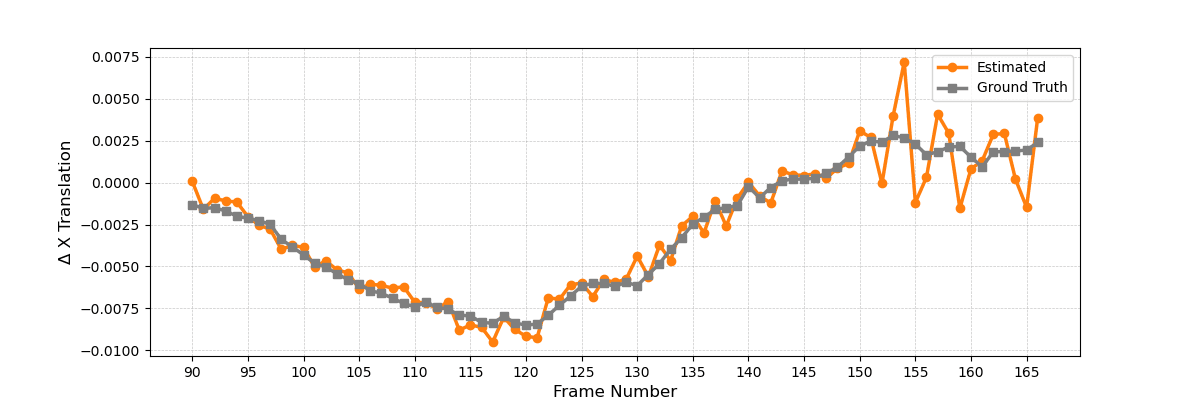}
        \caption*{14-03: $\Delta Y$ Translation}
    \end{subfigure}

    \vspace{1em}

    \begin{subfigure}[t]{0.49\linewidth}
        \centering
        \includegraphics[width=\linewidth]{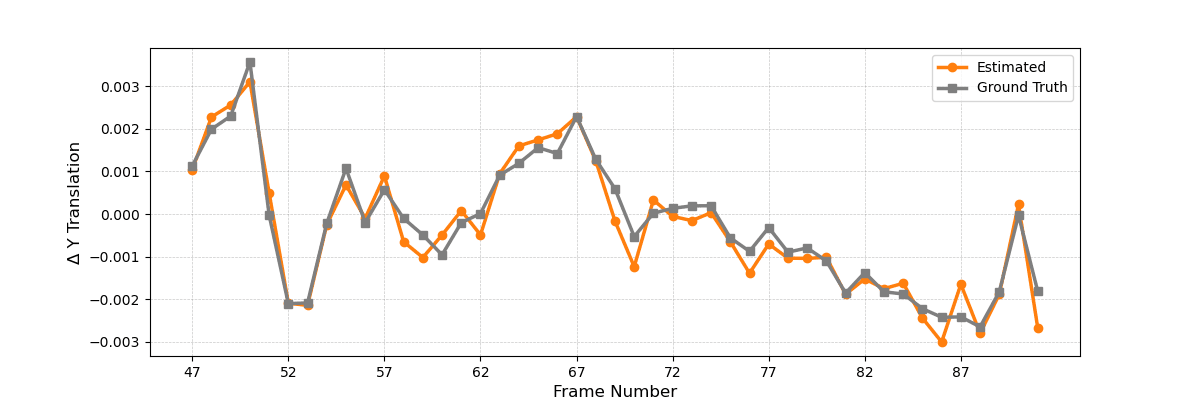}
        \caption*{14-04: $\Delta X$ Translation}
    \end{subfigure}
    \hfill
    \begin{subfigure}[t]{0.49\linewidth}
        \centering
        \includegraphics[width=\linewidth]{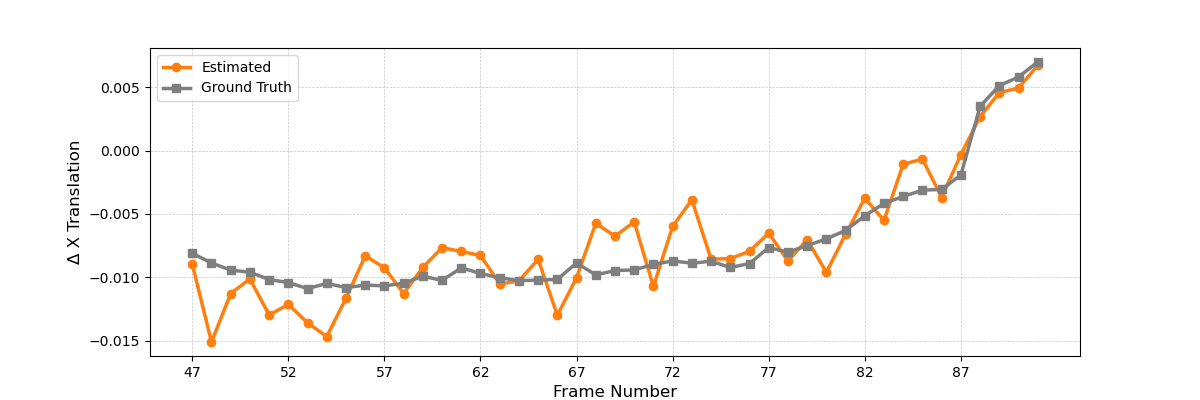}
        \caption*{14-04: $\Delta Y$ Translation}
    \end{subfigure}

    \caption{Comparison of estimated and ground truth object translation along horizontal ($\Delta X$) and vertical ($\Delta Y$) image axes for sequences 13-05, 14-03, and 14-04. in EVIMO2v2}
    \label{fig:translation_comparison_all}
\end{figure}

\section{Conclusion}
\label{sec:conclusion}

We presented a method for motion segmentation and egomotion estimation using event-based normal flow, relying on geometric constraints and temporal consistency. In the current work, we demonstrate that leveraging temporal motion information and normal flow enables highly efficient and robust segmentation without requiring explicit depth or full optical flow estimation. Our formulation allows us to isolate independently moving objects through residual analysis and merge motion-consistent regions using hierarchical clustering.

Looking ahead, we aim to incorporate deep learning to enhance object boundary separation and streamline the clustering process, thereby reducing reliance on fine-grained merging operations. Additionally, the translation velocities estimated from normal flow, together with temporal geometric cues, will further support fast and accurate assignment of background, foreground, and object identities—paving the way for scalable, real-time event-based perception systems.

{
    \small
    \bibliographystyle{ieeenat_fullname}
    \bibliography{main}
}

\end{document}